%% file: main.tex
\documentclass[10pt,letterpaper]{article}

\usepackage{times}
\usepackage{epsfig}
\usepackage{graphicx}
\usepackage{amsmath}
\usepackage{amssymb}

\usepackage{diagbox}
\usepackage{bbm}
\usepackage [english]{babel}
\usepackage [autostyle, english = american]{csquotes}
\usepackage[nottoc]{tocbibind}
\MakeOuterQuote{"}

\usepackage{multirow}
\usepackage{footmisc}
\usepackage{color}
\usepackage[dvipsnames]{xcolor}
\usepackage{colortbl,booktabs}
\usepackage{threeparttable}
\usepackage{nopageno}
\usepackage{comment}
\usepackage[small,bf]{caption}
\usepackage{url}
\usepackage{algorithm}
\usepackage{algpseudocode}
\usepackage{mathtools}
\usepackage{pifont}
\usepackage{lipsum}

\usepackage{url}

\usepackage{lipsum}
\usepackage{color}
\definecolor{Sijia_color}{rgb}{0.858, 0.188, 0.478}
\usepackage{wrapfig}
\usepackage{booktabs}
 \usepackage{multirow} 
 
\usepackage{diagbox}
\usepackage{bbm}

\usepackage{adjustbox}
\usepackage{multirow}
\usepackage{footmisc}
\usepackage{colortbl,booktabs}
\usepackage{threeparttable}
\usepackage{nopageno}
\usepackage{xspace}

\usepackage{wrapfig}
\input{math_commands.tex}

\DeclareMathOperator*{\minimize}{\text{minimize}}
\DeclareMathOperator*{\maximize}{\text{maximize}}

\DeclareMathAlphabet\mathbfcal{OMS}{cmsy}{b}{n}
\newcommand{\Def}[0]{\mathrel{\mathop:}=}

\def\affine{\textit{advT-Affine}\xspace}
\def\TPS{\textit{advT-TPS}\xspace}
\def\baseline{\textit{advPatch}\xspace}

\begin{document}
% \linenumbers
% \pagestyle{headings}
% \mainmatter
% \def\ECCVSubNumber{4383}  % Insert your submission number here

% \title{Adversarial T-shirt! \\Evading Person Detectors in A Physical World} % Replace with your title

%\begin{comment}
% \titlerunning{ECCV-20 submission ID \ECCVSubNumber} 
% \authorrunning{ECCV-20 submission ID \ECCVSubNumber} 
% \author{Anonymous ECCV submission}
% \institute{Paper ID \ECCVSubNumber}
%\end{comment}
%******************

% \begin{comment}
\title{Adversarial T-shirt! \\Evading Person Detectors in A Physical World}
\author{Kaidi Xu$^{1}$ \,  Gaoyuan Zhang$^{2}$  \, Sijia Liu$^{2}$  \, Quanfu Fan$^{2}$ \,  Mengshu Sun$^{1}$\\ \quad  {Hongge Chen}$^{3}$ \quad  
{Pin-Yu Chen}$^{2}$ \quad 
  {Yanzhi Wang}$^{1}$  \quad  {Xue Lin}$^{1}$ \\
\\ 
$^1$Northeastern University, USA  \\
$^2$MIT-IBM Watson AI Lab, IBM Research,  USA\\
$^3$Massachusetts Institute of Technology, USA
}

\maketitle

\begin{abstract}
It is known that deep neural networks (DNNs) are vulnerable to adversarial attacks. 
The so-called \textit{physical adversarial examples} deceive DNN-based decision makers by attaching adversarial patches to   real objects.
% Even in the real world,  by adding crafted perturbation on the object or background can lead deep neural networks misclassified. 
However, most of the existing works on physical adversarial attacks focus on static objects such as glass frames, stop signs and images attached to cardboard. In this work, we propose \textit{Adversarial T-shirts},   a robust physical adversarial example for evading person detectors even if it
%the adversarial T-shirt 
%\PY{I think we should highlight all current physical adversarial examples suffer from deformation, not ours suffer from it} 
%suffers from 
could undergo non-rigid deformation due to a moving person's pose changes. To the best of our knowledge, this is the first work that models the effect of deformation for designing  physical adversarial examples with respect to non-rigid objects such as T-shirts.
%using  thin plate spline (TPS) transformation. 
We show that the proposed method  achieves  74\%  and  \textcolor{black}{57\%}  attack  success  rates in the digital and physical worlds respectively against YOLOv2. In contrast, the state-of-the-art physical attack method to fool a person detector   only achieves
\textcolor{black}{18\%}
attack success rate.
Furthermore, by leveraging min-max optimization, we extend our method to the ensemble attack setting against two object detectors YOLO-v2  and Faster R-CNN simultaneously.
% \keywords{Physical adversarial attack; object detection; deep learning
% }
\end{abstract}

%%%%%%%%% BODY TEXT
\begin{figure*}[t!]
   \centering
\adjustbox{max width=1\textwidth}{
\hspace*{-0.1in}\begin{tabular}{p{0.22in}p{0.65in}p{0.65in}p{0.65in}p{0.65in}p{0.65in}p{0.65in}p{0.65in}}
& 
 \hspace*{-0.1in} \parbox{0.65in}{\centering \footnotesize frame 5} & \parbox{0.65in}{\centering \footnotesize frame 30}
& \parbox{0.65in}{\centering \footnotesize frame 60} & \parbox{0.65in}{\centering \footnotesize frame 90}
& \parbox{0.65in}{\centering \footnotesize frame 120} & \parbox{0.65in}{\centering \footnotesize frame 150}
& \parbox{0.65in}{\centering \footnotesize pattern}
\vspace{-10mm}
\\
%\rotatebox{90}{ \footnotesize \ \ \ \ Adversarial}
\hspace*{0.05in} \rotatebox{90}{\parbox{1.2in}{\centering \footnotesize  Ours (digital)   \\ adversarial T-shirt }}&\hspace*{-0.1in} 
\includegraphics[width=0.68in]{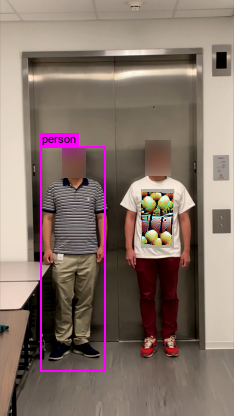}& \hspace*{-0.1in}  
\includegraphics[width=0.68in]{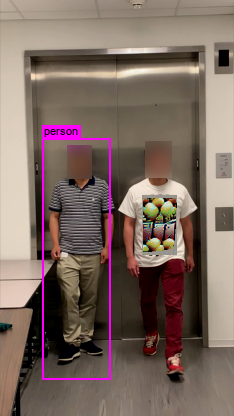}& \hspace*{-0.1in} \includegraphics[width=0.68in]{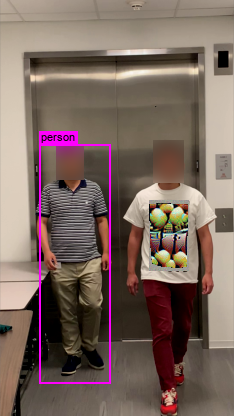}& \hspace*{-0.1in} \includegraphics[width=0.68in]{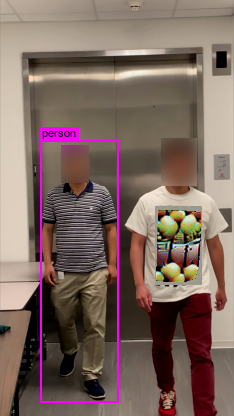}& \hspace*{-0.1in} \includegraphics[width=0.68in]{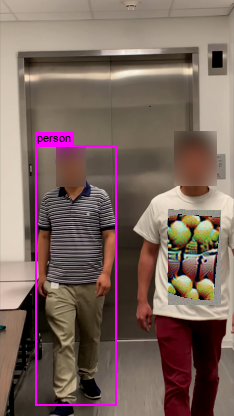}& \hspace*{-0.1in} \includegraphics[width=0.68in]{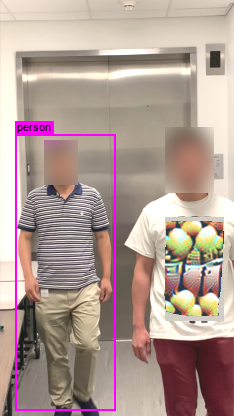}& \hspace*{-0.1in} 
\includegraphics[width=0.68in, height =1.2in]{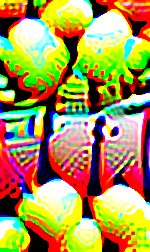}
\\ 
%  \begin{tabular}[l]{@{}l@{}}
 %\rotatebox{90}{ \footnotesize{Africa elephant}}  
 %\rotatebox{90}{\parbox{0.9in}{\centering Africa elephant}}
 \hspace*{0.05in} \rotatebox{90}{\parbox{1.2in}{\centering \footnotesize Ours (physical):\\Adversarial T-shirt}}
 &   \hspace*{-0.1in} 
\includegraphics[width=0.68in]{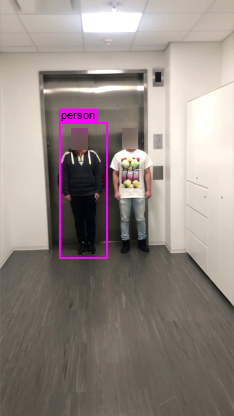}& \hspace*{-0.1in}  
\includegraphics[width=0.68in]{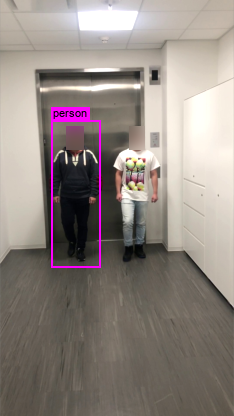}& \hspace*{-0.1in}  \includegraphics[width=0.68in]{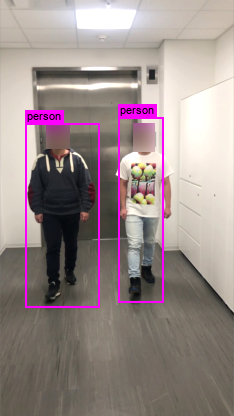}& \hspace*{-0.1in}  \includegraphics[width=0.68in]{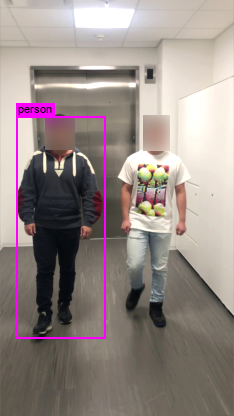}& \hspace*{-0.1in}  \includegraphics[width=0.68in]{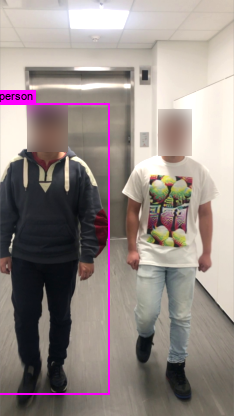}& \hspace*{-0.1in}  \includegraphics[width=0.68in]{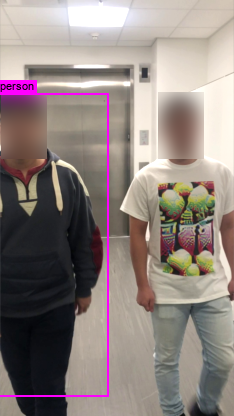}& \hspace*{-0.1in}  
\includegraphics[width=0.68in, height =1.2in]{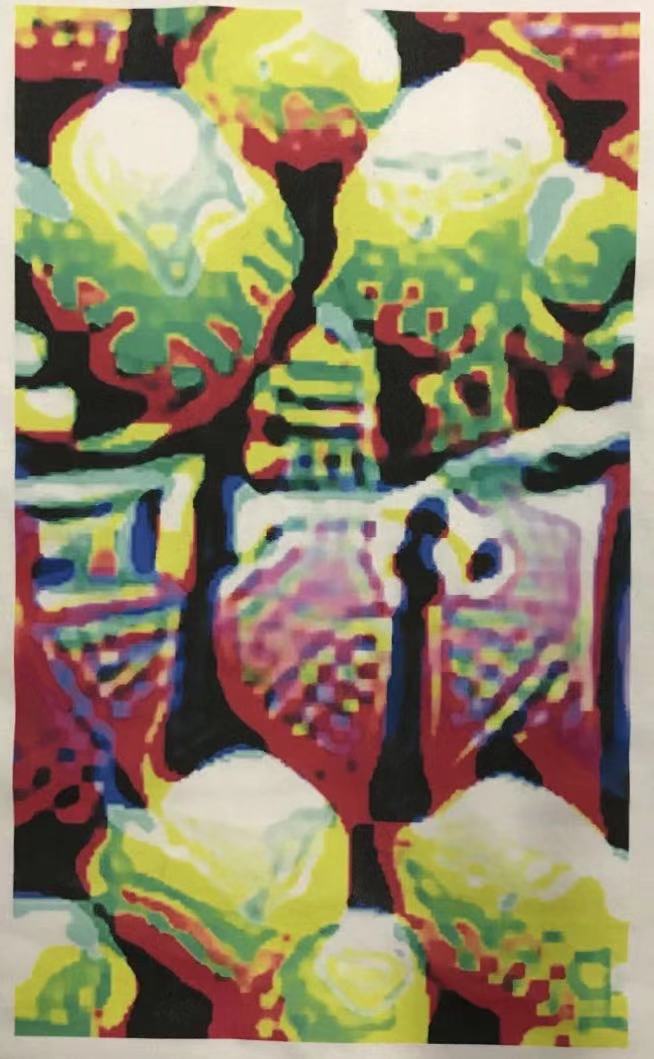}
\\ 

%\rotatebox{90}{ \footnotesize \ \ \ \ Adversarial}
\hspace*{0.05in} \rotatebox{90}{\parbox{1.2in}{\centering \footnotesize  Affine (non-TPS)   \\ adversarial T-shirt }}&\hspace*{-0.1in} 
\includegraphics[width=0.68in]{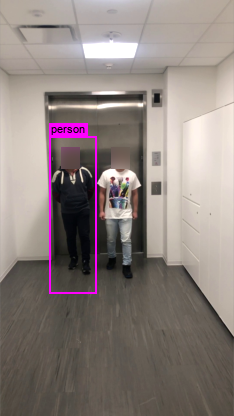}& \hspace*{-0.1in}  
\includegraphics[width=0.68in]{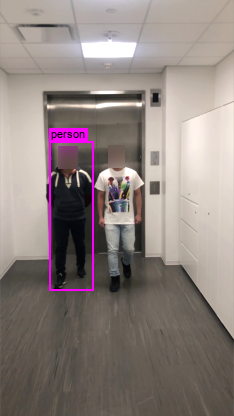}& \hspace*{-0.1in} \includegraphics[width=0.68in]{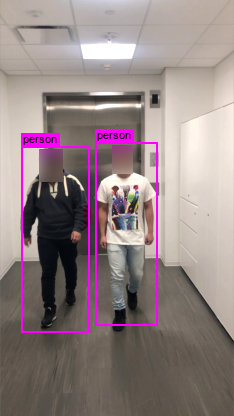}& \hspace*{-0.1in} \includegraphics[width=0.68in]{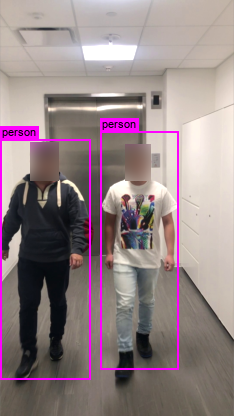}& \hspace*{-0.1in} \includegraphics[width=0.68in]{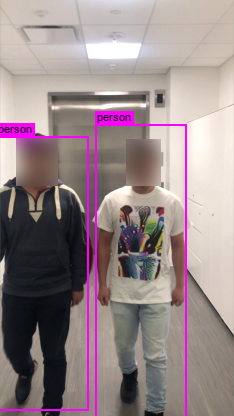}& \hspace*{-0.1in} \includegraphics[width=0.68in]{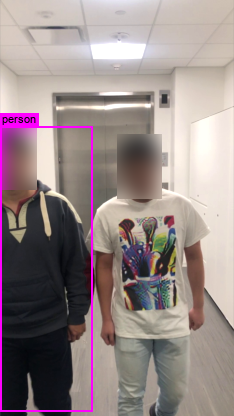}& \hspace*{-0.1in} 
\includegraphics[width=0.68in, height =1.2in]{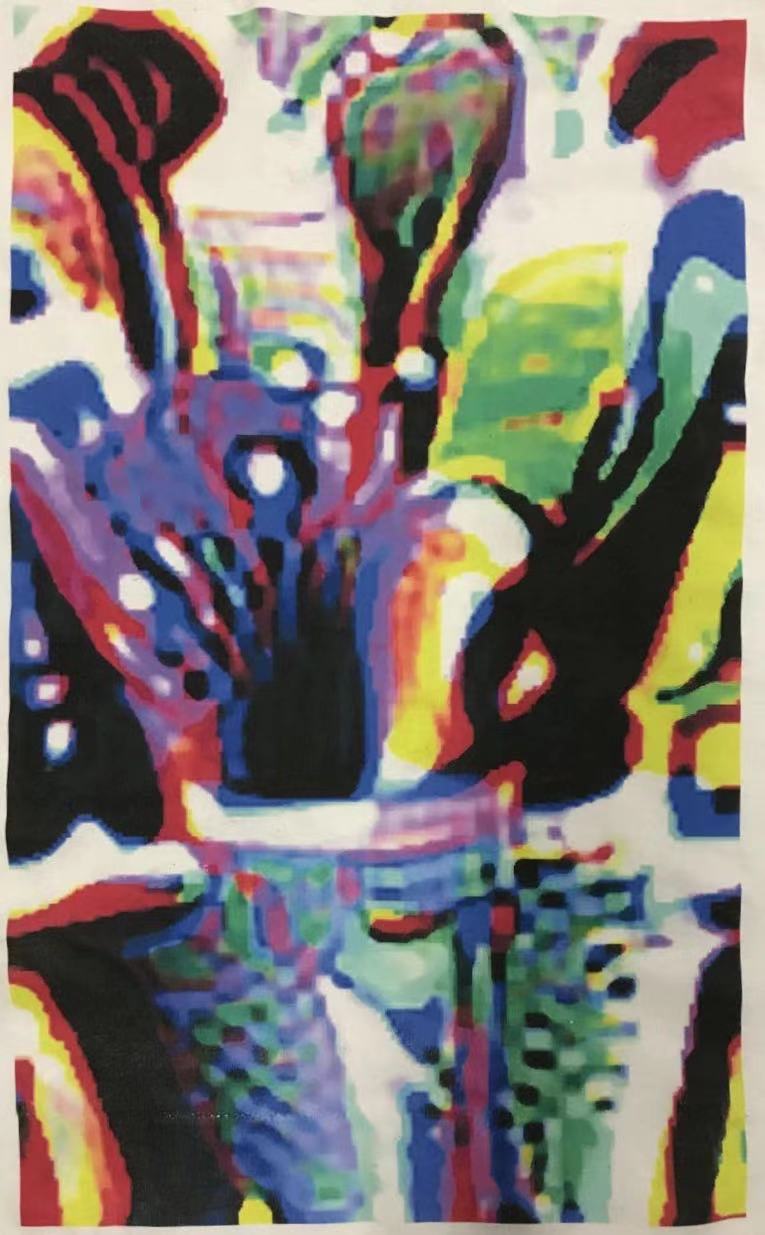}
\\ 
%\rotatebox{90}{ \footnotesize \ \ \ \ Adversarial}
\hspace*{0.05in} \rotatebox{90}{\parbox{1.2in}{\centering \footnotesize Baseline physical attack}}&\hspace*{-0.1in} 
\includegraphics[width=0.68in]{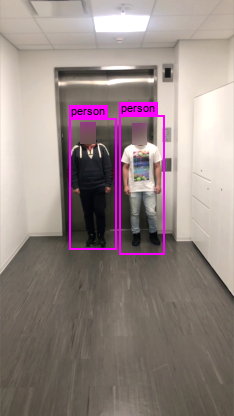}& \hspace*{-0.1in}  
\includegraphics[width=0.68in]{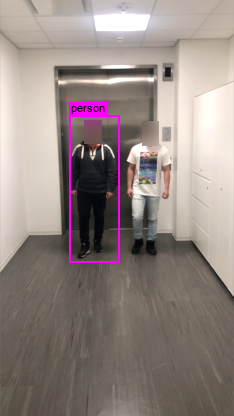}& \hspace*{-0.1in}  
\includegraphics[width=0.68in]{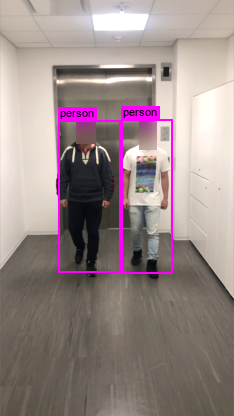}& \hspace*{-0.1in}  
\includegraphics[width=0.68in]{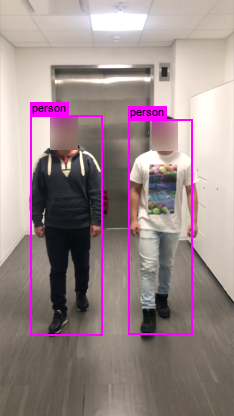}& \hspace*{-0.1in}  
\includegraphics[width=0.68in]{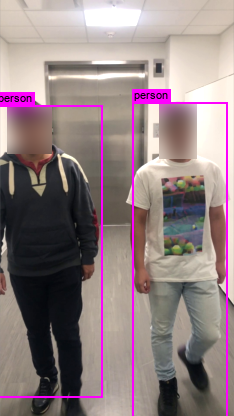}& \hspace*{-0.1in}  
\includegraphics[width=0.68in]{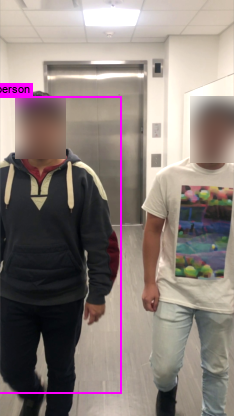}& \hspace*{-0.1in}  
\includegraphics[width=0.68in, height =1.2in]{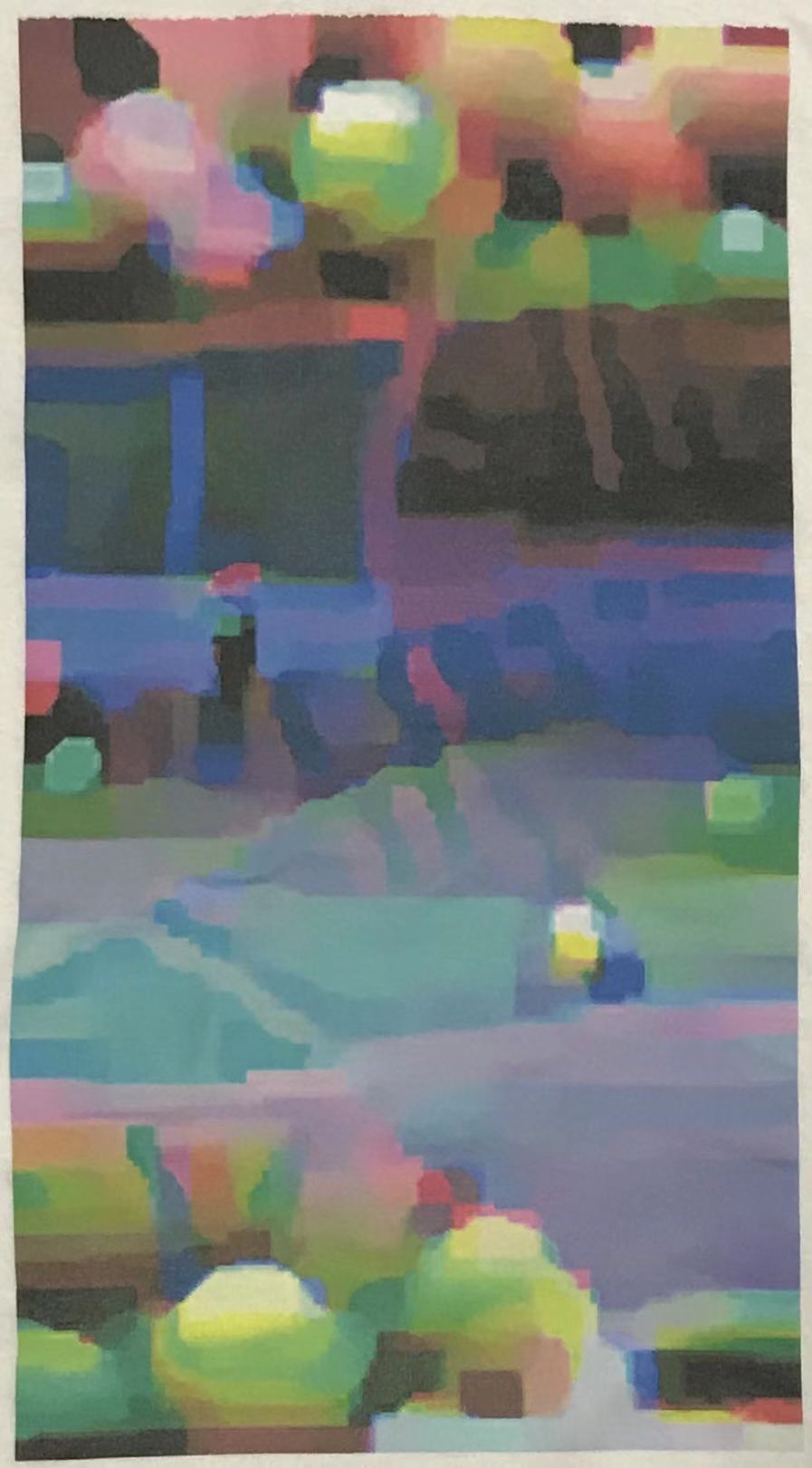}

    \end{tabular}}
\vspace{-5mm}
\caption{\footnotesize{Evaluation of the effectiveness of adversarial T-shirts to evade person detection by YOLOv2. Each row corresponds to a specific attack method while each column except the last one shows an individual frame in a video. The last column shows the adversarial patterns applied to the T-shirts. At each frame, there are two persons, one of whom wears the adversarial T-shirt. First row: digital adversarial T-shirt generated using TPS.
Second row: physical adversarial T-shirt generated using TPS. Third row: physical adversarial T-shirt generated using {affine transformation} (namely, in the absence of TPS). Fourth row: {T-shirt with physical adversarial patch considered in \cite{thys2019fooling}  to evade person detectors}. 
%\textcolor{Sijia_color}{[SL: will update caption.]}
}}
    \label{fig: intro}
    \vspace{-4mm}
\end{figure*}

\section{Introduction}

% \begin{itemize}
%     \item Background introduction. Starting from classifier to object detector, digital world to physical world. More and more challenge.
%     \item Existed methods and their weakness.
%     \item Our contribution: 1. TPS 2. General transformation library and framework. 3. Ensemble attack
% \end{itemize}

The vulnerability of deep neural networks (DNNs) against adversarial attacks (namely, perturbed inputs deceiving DNNs) has been found in   applications spanning from image classification to speech recognition \cite{goodfellow2014explaining,xu2019structured,zhao2019admm,carlini2018audio,xu2019topology,athalye2018obfuscated}. 
Early works  studied adversarial examples only in the digital space. Recently, some works showed that  it is possible to create  adversarial perturbations   on physical objects and fool DNN-based decision makers under a variety of real-world conditions \cite{sharif2016accessorize,eykholt2018robust,athalye18b,eykholt2018physical,lu2017adversarial,chen2018shapeshifter,thys2019fooling,cao2019adversarial,li2019adversarial}. {The design of   \textit{physical adversarial attacks} helps to evaluate the    robustness of  DNNs deployed in   real-life systems,  e.g., autonomous vehicles and    surveillance systems.}
However, most of the studied {physical adversarial attacks}   encounter two limitations: a)   the physical objects are usually considered  being \textit{static}, 
%(e.g., adversarial stop sign and glass frame), 
and b) the possible  \textit{deformation} of adversarial  pattern attached to   a moving object (e.g., due to pose change of a moving person) is commonly neglected. 
In this paper, we propose a new type of physical adversarial attack, \textit{adversarial T-shirt},  to evade  DNN-based person detectors  when a person wears the adversarial T-shirt; {see the second row of Fig.~\ref{fig: intro} for illustrative examples}. 
%\textcolor{Sijia_color}{[Put   figure1 at the first page. ]}
%\footnote{\KD{Different with the adversarial T-shirt sells on market, our \textit{adversarial T-shirt} is designed for evading person detectors.
%}}
% We will show that under this more realistic scenario, the existing physical adversarial patch  attached on
%   a cardboard \cite{thys2019fooling}   \textit{fails} to hide a person  (whom hold the cardboard) from surveillance cameras. By contrast, our   proposed \textit{adversarial T-shirt} achieves xxxx\%  and xxxx\%  attack success rates   in the digital  and the physical world, respectively.
%fooling an object detector fails, 
%  (e.g., adversarial perturbations on a paperboard worn  )
%the movement of the object could cause 
%solid rock undergoing slow deformation
% The lack of adversarial robustness for deep neural networks (DNNs) has brought in a significant amount of attention

\paragraph{Related work}
\textcolor{black}{Most of the existing    physical adversarial attacks are generated against  image classifiers and object detectors. %\textcolor{green}{[do we need cite here again? it totally same with the citations in  introduction]}.
In  \cite{sharif2016accessorize}, 
a face recognition system is fooled by a real eyeglass frame designed under a crafted   adversarial pattern. 
\textcolor{black}{In \cite{eykholt2018robust}, a stop sign is misclassified by adding black or white stickers on it against the image classification system.}
In \cite{li2019adversarial}, an image classifier is fooled by 
  placing a crafted sticker at the lens of a camera.
In \cite{athalye18b}, a so-called Expectation over Transformation (EoT)  framework was proposed to synthesize  adversarial examples robust to a set of physical transformations   such as  rotation, translation, contrast, brightness, and random noise. Compared to attacking image classifiers, generating physical adversarial attacks against object detectors is more involved. For example, the adversary 
is required to 
% \textcolor{Sijia_color}{[SL: This is an issue. @Kaidi, we need to clarify the attacking objective, and the relationship between bounding box detector and object classification.]}
mislead the  bounding box detector of an object   {when attacking YOLOv2~\cite{redmon2017yolo9000} and SSD~\cite{liu2016ssd}}. 
% \QF{it seems to me that misleading either the bounding box detector or the object classifier is ok? } 
A well-known  success of such attacks in the physical world is the generation of adversarial stop sign \cite{eykholt2018physical}, which deceives state-of-the-art object detectors such as YOLOv2  and Faster R-CNN  \cite{ren2015faster}. 
% \textcolor{black}{The work \cite{eykholt2018physical}, extended \cite{eykholt2018robust} to object detectors and attack stop sign by adding colorful stickers on it. }
%  intruders can sneak around
% undetected by holding a small cardboard plate in front of
% their body aimed towards the surveilance camera.
}

The most relevant approach to ours is the work of  \cite{thys2019fooling}, which demonstrates that a  person can evade a detector by holding a cardboard with an adversarial patch.
%  makes intruders   
% undetected by a  person detector when they 
% hold  the    adversarial patch attached on  a   cardboard plate.
However, such a physical attack restricts the adversarial patch to be attached to   a   \textit{rigid} carrier (namely, cardboard), 
%{and is not directly applied to the design of adversarial T-shirt.} 
{and is different from our setting here where the generated adversarial pattern   %from our approach 
is directly printed on a T-shirt.}
We   show that
the attack proposed by \cite{thys2019fooling} becomes ineffective  when the adversarial patch is   attached to a T-shirt (rather than a cardboard) and  worn by a moving person {(see the fourth row of Fig.~\ref{fig: intro})}.
{At the technical side, different from \cite{thys2019fooling} we propose  a    
thin plate spline (TPS) based transformer to model deformation of non-rigid objects, and develop an ensemble physical attack that fools object detectors YOLOv2 and Faster R-CNN simultaneously.
{We highlight that our proposed adversarial T-shirt is not just a T-shirt with printed adversarial patch for clothing fashion,  it is a physical adversarial wearable designed for  evading person detectors in the real world. 
% the proposed adversarial T-shirt
% is not just 
% Different with the adversarial T-shirt sells on market, our \textit{adversarial T-shirt} is designed for evading person detectors.
}
 }

\textcolor{black}{
Our work is also motivated by the importance of person detection on  intelligent surveillance.  DNN-based 
    surveillance systems    have significantly advanced the field of object detection \textcolor{black}{\cite{girshick2014rich,girshick2015fast}}.
 Efficient   object detectors such as faster R-CNN \cite{ren2015faster},  SSD  \cite{liu2016ssd}, and YOLOv2 \cite{redmon2017yolo9000}  have been  deployed for human detection.
 Thus,   one may wonder whether or not there exists 
 a    security risk for intelligent surveillance systems caused by  adversarial human wearables, e.g., adversarial T-shirts.  
However, paralyzing a person detector in the physical world   requires substantially more  challenges such as low resolution, pose changes and occlusions. {The success of our adversarial T-shirt against real-time person detectors offers new insights for designing practical physical-world adversarial human wearables. }
}

\textcolor{black}{
\paragraph{Contributions}\textcolor{black}
{We summarize our contributions as follows:}
\begin{itemize}
    \item We develop a   
  TPS-based transformer to model the temporal deformation of an adversarial T-shirt caused by pose changes of a moving person. We also show the importance of such non-rigid transformation to ensuring  the effectiveness of adversarial T-shirts in the physical world. 
  \item We propose a general optimization framework for design of adversarial T-shirts in both single-detector and multiple-detector settings.
  \item We conduct experiments in both digital and physical worlds and show that the proposed adversarial T-shirt achieves \textcolor{black}{74\%  and 57\%}  attack success rates respectively when attacking YOLOv2. By contrast, {the physical adversarial patch~\cite{thys2019fooling} printed on a T-shirt} only achieves \textcolor{black}{18\%} attack success rate. Some of our results are highlighted in Fig.\,\ref{fig: intro}.
  %\footnote{All images including real person are face blurred (after detection).} 
%   \PY{These numbers are different from Abtract}
\end{itemize}
}

\section{Modeling Deformation of A Moving Object by Thin Plate Spline Mapping}
\label{sec: transformer}

\textcolor{black}{In this section, we begin by reviewing some existing transformations required in the design of physical adversarial examples. We then elaborate on the Thin Plate Spline (TPS) mapping we adopt in this work to model the possible deformation encountered by a moving and non-rigid object.}

% To design physical adversarial examples, a set of transformations are often used to approximate a realistic space of possible distortions involved in printing out an
% image and taking a natural picture of it.
Let $\mathbf x$ be an original image (or a video frame), and $t(\cdot)$ be the physical transformer. The transformed image $\mathbf z$  under $t$ is given by
\begin{align}\label{eq: transform_x}
\mathbf z =     t(\mathbf x).
\end{align}
%where $\boldsymbol{\theta}$ are parameters associated with the transformer $t$.

%\subsection{Existing transformers}
%\subsection{\textcolor{Sijia_color}{Existing transformations and their limitations}}

\paragraph{\textcolor{black}{Existing transformations.}}
In \cite{athalye18b}, the parametric transformers include  scaling, translation, rotation, brightness and additive Gaussian noise;  see details in \cite[Appendix\,D]{athalye18b}.
% set of random transformations of the form $t(\mathbf x) = \mathbf A \mathbf x + \mathbf b$ is considered. Given a
% particular choice of transformer parameters $\boldsymbol{\theta}$, the rendering of $\mathbf x$ 
% can be written as $\mathbf A \mathbf x + \mathbf b$ for some coordinate map $\mathbf A$ and
% background $\mathbf b$. The used transformations include \textit{scale, rotation, lighten/darken, Gaussian noise, and translation}; see details in \cite[Appendix\,D]{athalye18b}.
In \cite{liu2018beyond},  the geometry and lighting transformations are studied via parametric models.
Other transformations including perspective transformation, brightness adjustment,  resampling (or image resizing), smoothing and saturation are considered in   \cite{sitawarin2018rogue,ding2019sensitivity}.  
\textcolor{black}{All the existing transformations  are included in our library of physical transformations.}
However, they  are not sufficient to model the cloth deformation 
caused by pose change of a moving person. \textcolor{black}{
For example,  the {second and third} rows of Fig.\,\ref{fig: intro} show that
adversarial T-shirts designed  against only  existing physical transformations yield low attack success rates.  
%compared to our proposed adversarial T-shirt by taking into account the effect of cloth deformation introduced in what follows.
} 
%\textcolor{Sijia_color}{[SL: Needs a figure to show the digital world example of adversarial T-shirt using existing transformations almost fails in the testing video but succeeds when TPS is considered.]}

% \textcolor{Sijia_color}{\paragraph{The need of non-rigid transformations}[Needs one figure (e.g., additional row of  Fig. 1 to show attack without  TPS fails even in the digital world.) which shows that existing transformations are not sufficient to render robustness of adversarial T-shirt in physical environment.]}

% \textcolor{Sijia_color}{SL: Please understand how to generate these transformations in practice. Is there an explicit form for those transformers?}

%\subsection{Non-rigid transformation via  thin plate spline (TPS) mapping}
%Motivation
%Learning physical adversarial patterns effectively is a challenging problem as the image perturbations need to be robust under different camera views and lighting conditions. 

\begin{figure*}[htb]
   \centering
   \adjustbox{max width=1\textwidth}{
\hspace*{-0.05in}\begin{tabular}{cccc}
\includegraphics[width=1.17in]{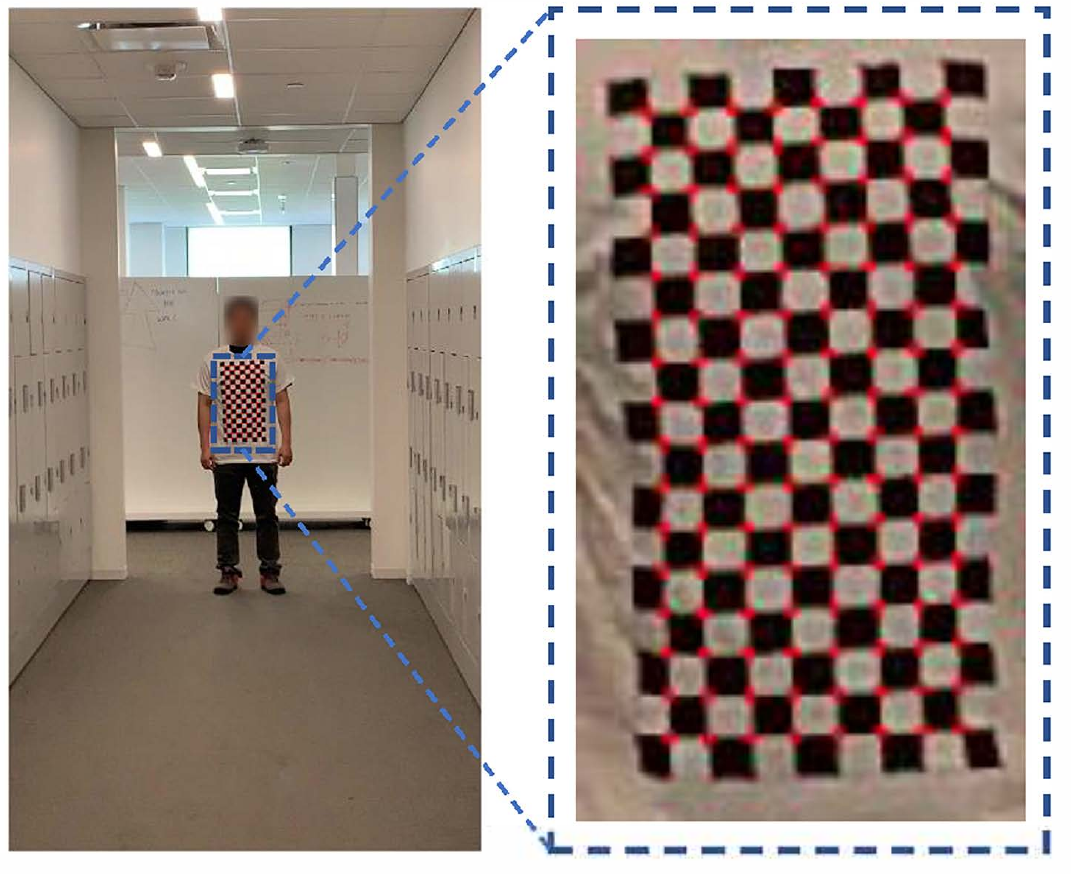}& 
\includegraphics[width=1.19in]{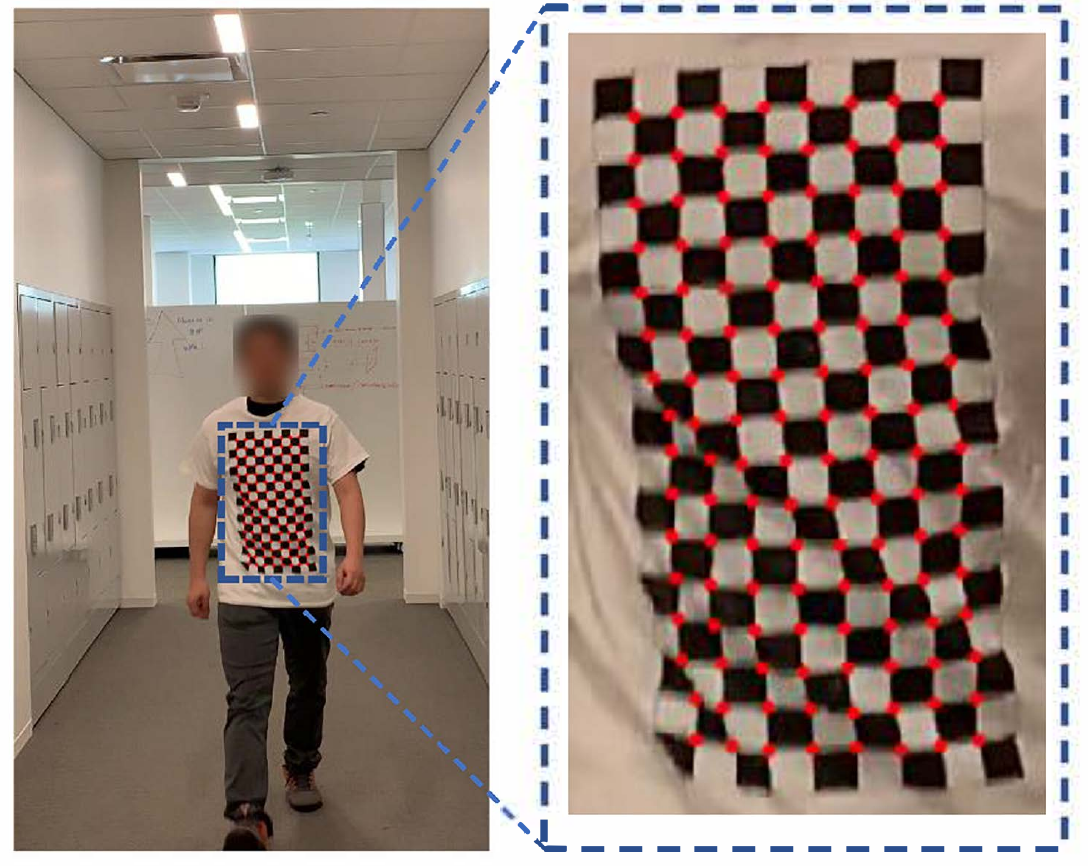}&
\includegraphics[width=1.05in, height = .92in]{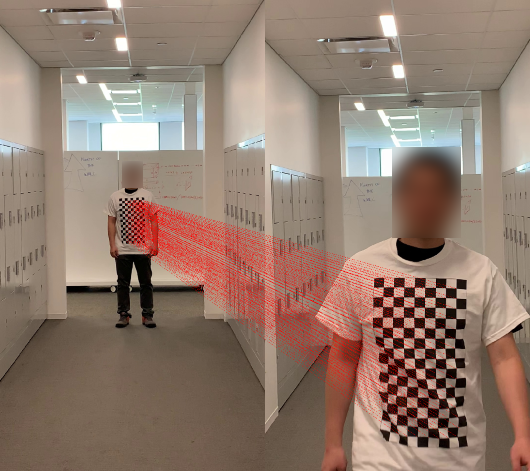}&
\includegraphics[width=1.23in]{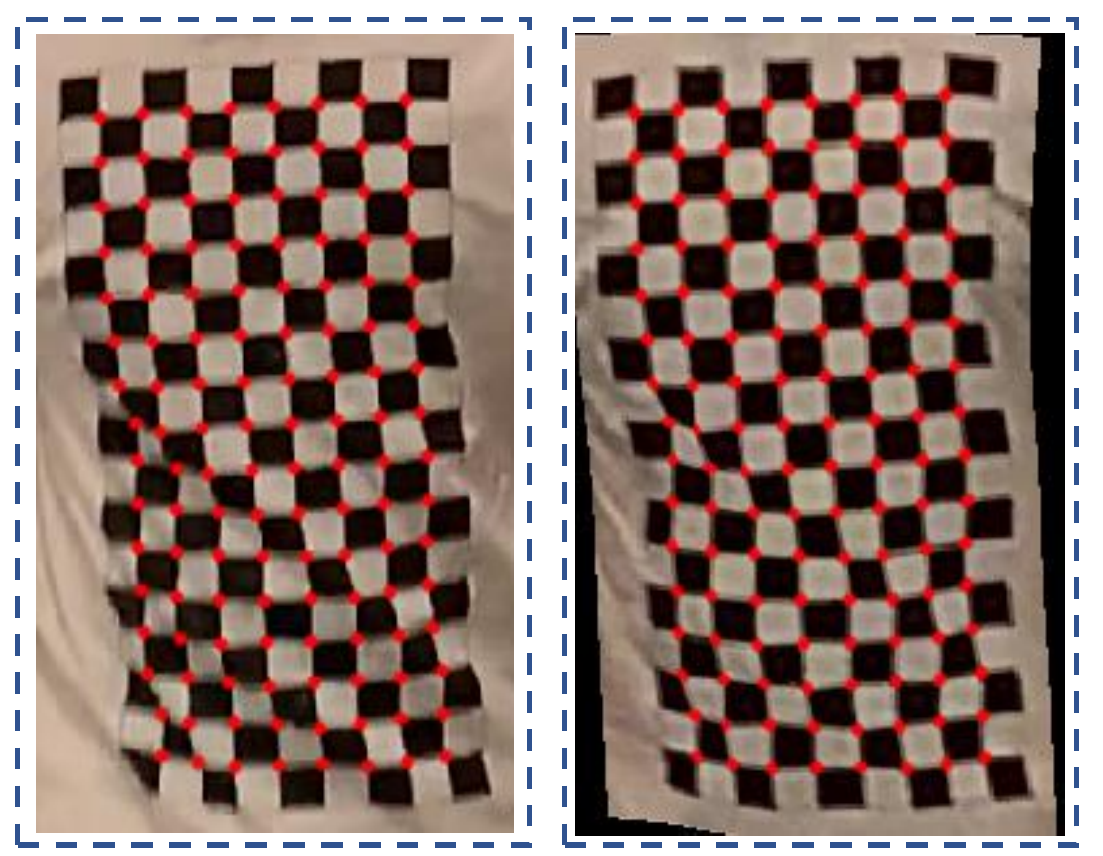}
\\   
(a) & (b)&(c) & (d)
\end{tabular}
}
\caption{\footnotesize{Generation of TPS. 
(a) and (b): Two frames with checkerboard detection results. (c): Anchor point matching process between two frames (d): Real-world close deformation in (b)  versus the synthesized TPS transformation (right plot).
}}
    \label{fig: tps_intro}
    %\vspace*{-4mm}
\end{figure*}

\paragraph{TPS transformation for cloth deformation.} %\textcolor{Sijia_color}{[I rewrite the following paragraphs. Please check. @Quanfu @Kaidi.]}
%Humans are non-rigid objects. 
A person's movement can result in significantly and constantly changing wrinkles (aka deformations) in her clothes. 
This makes it challenging to develop an adversarial T-shirt effectively in the real world. 
%since commonly-used linear models such as projective and affine transformations cannot capture such large deformations of the clothes. 
To circumvent this challenge, we employ TPS mapping  \cite{bookstein1989principal} to model the cloth deformation caused by human body movement. TPS has been widely used as the non-rigid transformation model in image alignment and shape matching~\cite{jaderberg2015spatial}. It consists of an {affine} component and a {non-affine} warping component. We will show that the non-linear warping part in TPS can provide an effective means of modeling cloth  deformation for learning adversarial patterns of non-rigid objects. %In addition, TPS produces smooth and infinitely differentiable surface for interpolation, ...... 
%We first give a brief review of TPS below.

%Thin plate Spline
%\textbf{Thin Plate Spline Interpolation.} 
 TPS learns a parametric deformation mapping from an original image $\mathbf x$ to a target image $\mathbf z$  through a set of control points with given positions. Let $\mathbf p \Def ( \phi, \psi ) $ denote  the 2D location of  an image pixel.  The deformation from $\mathbf x$ to $\mathbf z$ is then characterized by the \textit{displacement} of every pixel, namely,  how a pixel at $\mathbf p^{(x)} $ on image $\mathbf x$  changes to the pixel on image $\mathbf z$ at  $\mathbf p^{(z)} $, where $\phi^{(z)} = \phi^{(x)} + \Delta_{\phi} $ and $\psi^{(z)} = \psi^{(x)} + \Delta_{\psi} $, and $\Delta_{\phi}$ and $\Delta_{\psi}$ denote the pixel displacement  on image $\mathbf x$ along $\phi$ direction and $\psi$ direction, respectively. 
 
 Given a set of $n$ control points with locations $ \{ \hat{\mathbf p}_i^{( x)} \Def ( \hat \phi_{i}^{(x)}, \hat \psi_{i}^{(x)} ) \}_{i=1}^n$ on image $\mathbf x$, TPS provides a parametric model of pixel displacement when mapping     $\mathbf p^{(x)} $ to $\mathbf p^{(z)} $ \cite{chui2001non}
 %and locations 
% $ \{ \hat{\mathbf p}_i^{(z)} \Def  ( \hat \phi_{i}^{(z)}, \hat \psi_{i}^{(z)} ) \}_{i=1}^n$ on image $\mathbf z$, 
%  Let $\mathcal P_x \Def  \{ ( \phi_{x,i}, \psi_{x,i} ) \}_{i=1}^n$
%  and $\mathcal P_z \Def  \{ ( \phi_{z,i}, \psi_{z,i} ) \}_{i=1}^n$ denote the set of locations of anchor points in the image $\mathbf x$ and $\mathbf z$, respectively. Here $\mathbf p_i \Def [\phi_i, \psi_i]^T $ denotes the location of  pixel $i$ in an image. For each pixel  of image $\mathbf x$ at $(\phi_x, \spi_x)$, TPS 
%  $\mathbf{\hat{P}}=\{\hat{p}_i=(\hat{x}_i,\hat{y}_i) \in R^2 |i=\hdots n\}$ be a set of $n$ control points and $\mathbf{\hat{V}}=\{\hat{v}_i \in R |i=\hdots n\}$ be their corresponding function values. TPS interpolates the value of any 2D point $p=(x,y)$ by
\begin{align}\label{eq: TPS_form}
    \Delta(\mathbf p^{(x)}; \boldsymbol{\theta}) = & a_0 + a_1 \phi^{(x)} + a_2 \psi^{(x)}  + \sum_{i=1}^n c_i U (\| \hat{\mathbf p}_i^{( x)}  - \mathbf p^{(x)} \|_2),
\end{align}
where $U(r) = r^2 \log(r)$ and $\boldsymbol{\theta} = [\mathbf c; \mathbf a]$ are the TPS parameters, and $\Delta(\mathbf p^{(x)}; \boldsymbol{\theta})$ represents the displacement along either $\phi$ or $\psi$ direction. 
% The spline function above is composed of an affine transformation parameterized by $\mathbf a$ and non-affine warping specified by $\mathbf c$. 
% The affine part deals with transformations such as scaling and rotation while the second warping part models large deformations in clothes caused by person movement.

Moreover,  given the locations of control points on the transformed image $\mathbf z$ (namely, $ \{ \hat{\mathbf p}_i^{( z)}  \}_{i=1}^n$), 
TPS resorts to a regression problem to determine the parameters  $\boldsymbol{\theta}$ in \eqref{eq: TPS_form}.
% , TPS resorts to a regression model given the locations of control points on the transformed image $\mathbf z$, namely, $ \{ \hat{\mathbf p}_i^{( z)}  \}_{i=1}^n$. 
The regression objective is to minimize  the distance between $\{ \Delta_{\phi}(\mathbf p_i^{(x)}; \boldsymbol{\theta}_{\phi}) \}_{i=1}^n $  and $ \{ \hat {\Delta}_{\phi,i} \Def \hat \phi_{i}^{(z)} - \hat \phi_{i}^{(x)} \}_{i=1}^n$ along the $\phi$ direction,  
and   the distance between  $\{ \Delta_{\psi}(\mathbf p_i^{(x)}; \boldsymbol{\theta}_{\psi}) \}_{i=1}^n $  and $ \{  \hat {\Delta}_{\psi,i} \Def \hat \psi_{i}^{(z)} - \hat \psi_{i}^{(x)} \}_{i=1}^n$ along the $\psi$ direction, respectively. Thus,    TPS \eqref{eq: TPS_form} is applied to    coordinate $\phi$ and $\psi$ separately (corresponding to parameters $\boldsymbol{\theta}_{\phi}$ and $\boldsymbol{\theta}_{\psi}$). 
% To determine the parameters $\boldsymbol{\theta}$, it is easy to think of the control points as position constraints on a bending surface. Thus we minimize the bending energy of $f$ and the fitting error given below,
% \begin{align}
% E_{tps}(f)=\sum_{i=1}^{i=n} ||\hat{v}_i-f(\hat{x}_i,\hat{y}_i)||^2+\lambda\int\int_{R^2}(f^2_{xx}+2f^2_{xy}+f^2_{yy})dxdy
% \end{align}
% where $\lambda$ is a predefined coefficient that controls the significance of the bending energy item in the optimization.
% This leads to a
The regression problem   can be solved by the following linear system of equations~\cite{donato2002approximate}
\begin{align}\label{eq: TPSx}
\begin{bmatrix}
\mathbf  K &  \mathbf  P \\
\mathbf P^T & \mathbf 0_{3 \times 3}
\end{bmatrix} 
\boldsymbol{\theta}_{\phi}  =
\begin{bmatrix}
\hat{\boldsymbol  \Delta}_\phi  \\
\mathbf 0_{3 \times 1}
\end{bmatrix}, ~
\begin{bmatrix}
\mathbf  K &  \mathbf  P \\
\mathbf P^T & \mathbf 0_{3 \times 3}
\end{bmatrix} 
\boldsymbol{\theta}_{\psi}  =
\begin{bmatrix}
\hat{\boldsymbol  \Delta}_\psi  \\
\mathbf 0_{3 \times 1}
\end{bmatrix},
\end{align}
where the $(i,j)$th element of $\mathbf K \in \mathbb R^{n \times n}$ is  given by
$K_{ij}=U(\| \hat{\mathbf p}_i^{(x)}  - \hat{\mathbf p}_j^{(x)} \|_2)$, the $i$th row of $\mathbf P \in \mathbb R^{n \times 3}$ is given by $P_i = [1, \hat{\phi}_i^{(x)},\hat{\psi}_i^{(x)}]$, and the $i$th elements of $\hat{\boldsymbol  \Delta}_\phi \in \mathbb R^n$ and $\hat{\boldsymbol  \Delta}_\psi \in \mathbb R^n$ are given by $\hat{\Delta}_{\phi,i}$ and $\hat{\Delta}_{\psi,i}$, respectively. 

% To warp an image to another one with TPS, we need to compute the displacement of every pixel in the source image to the target image. %Let $\^{P}_s$ be a set of control points on the source image and $\^{P}_t$ their corresponding matching points be on the target image. 
% Let $dp=p_s-p_t=[dx, dy]^{T}$ be the displacement of a point $p_s$ on the source image and $p_t$ be the matching point of $p_s$ on the target image. Then two spline functions can be fit to compute $dx$ and $dy$ separately assuming that $dx$ is independent of $dy$.

\paragraph{Non-trivial application of TPS}
The   difficulty of implementing TPS for design of adversarial T-shirts exists from two aspects: 1) How to  determine the set of control points? And 2) how to   obtain  positions $\{ \hat{\mathbf p}_i^{(x)}\}$ and $\{ \hat{\mathbf p}_i^{(z)}\}$ of control points aligned between a pair of video frames $\mathbf x$ and $\mathbf z$?

% is  how to determine the set of control points and obtain positions $\{ \hat{\mathbf p}_i^{(x)}\}$ and $\{ \hat{\mathbf p}_i^{(z)}\}$ in both original and target images. 
\textcolor{black}{To address the first question,
we print  a \textit{checkerboard} on a T-shirt and use the camera calibration algorithm  \cite{geiger2012automatic,zhang2000flexible}  to detect points at the  intersection  between every two   checkerboard grid regions. These successfully detected points are considered as the  control points of one frame.   
Fig.\,\ref{fig: tps_intro}-(a) shows the checkerboard-printed T-shirt, together with the detected intersection points. 
Since TPS requires a set  of control points  \textit{aligned}  between  two frames, the second question on point matching arises. The challenge lies in the fact that   the control points detected at one video frame are different from those at another video frame (e.g., due to missing detection). Fig.\,\ref{fig: tps_intro}-(a) v.s. (b) provides an example of point mismatch. To address this issue, we adopt a 2-stage procedure, \textit{coordinate system alignment} followed by \textit{point aliment},  where the former refers to    conducting a perspective transformation from one frame to the other, and the latter finds the matched points at two frames through the nearest-neighbor method. We provide an illustrative example in Fig.\,\ref{fig: tps_intro}-(c). We refer readers to Appendix\,\ref{app: TPS} for more details about our   method. 
%of conducting a perspective transformation between two video frames.
}

\section{Generation of Adversarial T-shirt: An Optimization Perspective 
}\label{sec:tech}

%\textcolor{red}{[SL: leave some comments in this section. Please see Red texts.]}

In this section, we begin by formalizing the problem of adversarial T-shirt and introducing notations used in our setup.
We then propose to design a \textit{universal} perturbation used in our adversarial T-shirt to deceive a \textit{single} object detector.  %(e.g., YOLO or Faster R-CNN) over a \textit{sequence} of video frames. 
We lastly  propose a min-max (robust) optimization framework to design the universal adversarial patch against \textit{multiple} object detectors. 

\begin{figure*}[htb]
   \centering
\includegraphics[width=0.9\textwidth]{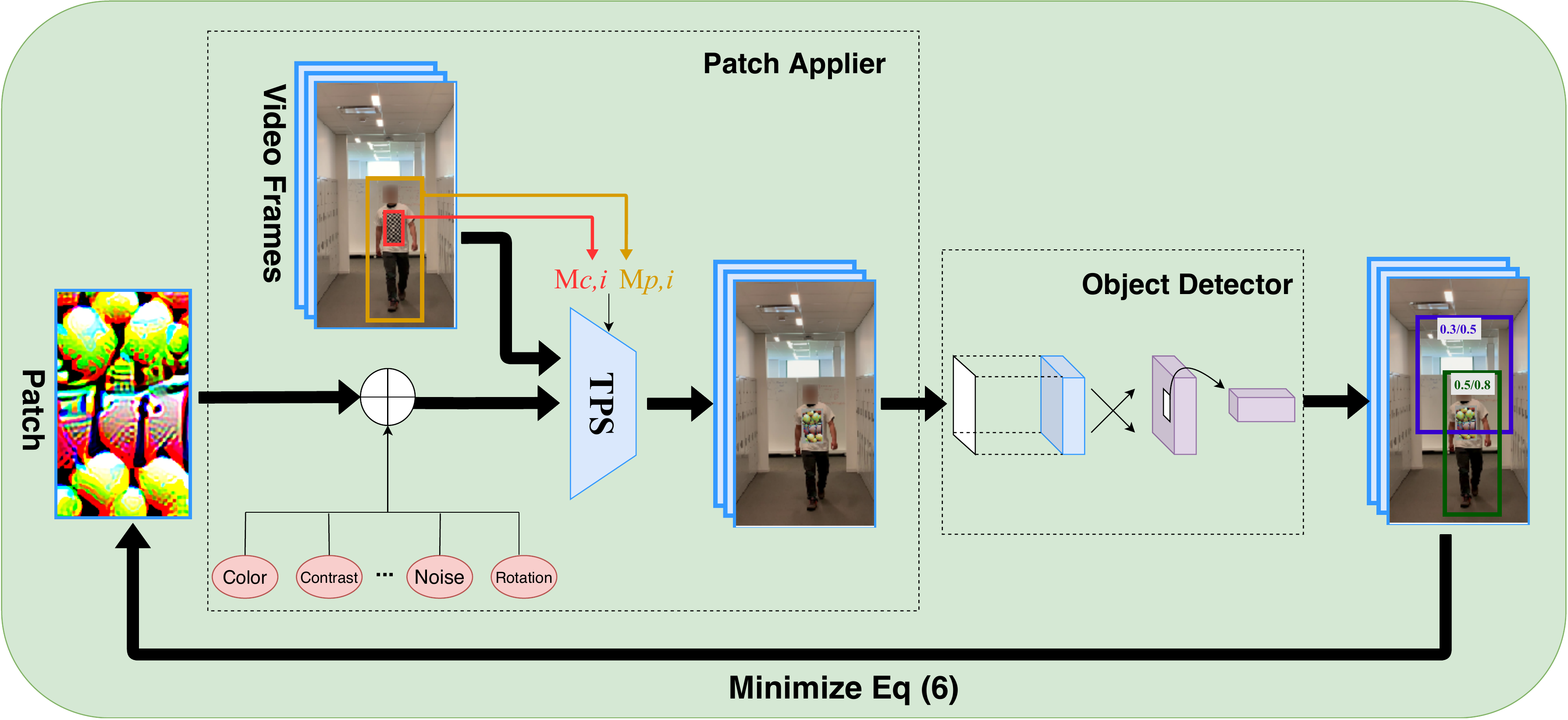}
\caption{
\textcolor{black}{Overview of the pipeline to generate   adversarial T-shirts. First, the  video frames containing a person whom wears the  T-shirt with printed   checkerboard pattern are used as training data.
Second, the universal adversarial perturbation (to be designed) applies to the cloth region by taking into account different kinds of transformations. Third, the adversarial perturbation is optimized through problem \eqref{eq: attack_single} by minimizing  the  largest bounding-box probability  belonging to the ‘person’ class.  The optimization procedure is performed as a closed loop through back-propagation. 
} 
}
    \label{fig: flowchart}
    %\vspace*{-4mm}
\end{figure*}

%\paragraph{Setup and notations.}
Let $\mathcal D \Def \{ \mathbf x_i \}_{i=1}^M$ denote $M$  video frames extracted from one or multiple given videos, where   $\mathbf x_i \in \mathbb R^d$ denotes the $i$th frame. 
%Let $\mathcal T  $ denote the set of conventional physical transformations, e.g., scaling, translation, rotation,   brightness and contrast. 
%Here   $t_i$ maps  the original image $\mathbf x$ to a transformed image $t_i(\mathbf x)$, which models the uncertainty of physical conditions. 
Let  $\boldsymbol{\delta} \in \mathbb R^d$  denote the universal adversarial perturbation applied to $\mathcal D$. 
The adversarial T-shirt is then characterized by $M_{c,i} \circ \boldsymbol{\delta}$, where $M_{c,i} \in \{ 0,1\}^d$  is a bounding box encoding the position  of the cloth region to be perturbed at the $i$th frame, and $\circ$ denotes element-wise product. 
\textit{The goal of adversarial T-shirt is to design $\boldsymbol{\delta}$ such that the perturbed frames of $\mathcal D$ are mis-detected by object detectors. % and are robust against physical transformations.
}
% \textcolor{red}{[Xue: $M_c$ is changing from frame to frame, how can we set one $M_c$ for the whole video?]}

\paragraph{Fooling a single object detector.} We generalize the Expectation over Transformation (EoT) method in \cite{athalye2018synthesizing} for design of adversarial T-shirts.
% It was shown in \cite{athalye2018synthesizing} that EoT provides a means to generate a robust adversarial perturbation to fool image classifiers in a physical world. However, 
Note that different from the conventional EoT,  a {transformers' composition}  is required for generating an adversarial T-shirt.
% the design of adversarial T-shirt since the latter to fool an object detector is built on  a \textit{hierarchical} transformation system.
For example, a perspective transformation on the  bounding box of  the T-shirt is composited with an TPS transformation applied to the   cloth region. 
%Moreover, in order to attach a universal adversarial perturbation to the T-shirt on a moving person, we also need to model the temporal change in the bounding box of both person and   T-shirt.
% \textcolor{red}{[Xue: I couldn't parse the last two sentences.]}

Let us begin by considering two video frames, an anchor image $\mathbf x_0$ (e.g., the first frame in the video) and a target image $\mathbf x_i$ for $i \in [M]$\footnote{$[M]$ denotes the integer set $\{ 1,2,\ldots, M\}$.}. 
Given the bounding boxes  of the person ($M_{p,0} \in \{ 0,1\}^d$) and the T-shirt ($M_{c,0} \in \{ 0,1\}^d$) at $\mathbf x_0$, we apply the perspective transformation  from $\mathbf x_0$ to $\mathbf x_i$ to obtain the   bounding boxes $M_{p,i}$ and $M_{c,i}$ at image $\mathbf x_i$. 
In the \textit{absence} of   physical transformations, the perturbed image $\mathbf x_i^\prime$ with respect to (w.r.t.) $\mathbf x_i$ is given by
\begin{align}\label{eq: perturbation_plain}
    \mathbf x_i^\prime =   \underbrace{{(\mathbf 1 - M_{p,i}) \circ \mathbf x_i}}_{\text{A}}  + \underbrace{M_{p,i} \circ \mathbf x_i }_{\text{B}}- \underbrace{M_{c,i} \circ \mathbf x_i}_{\text{C}} + \underbrace{M_{c,i} \circ \boldsymbol{\delta}}_{\text{D}},
\end{align}
where the term $A$ denotes the background region outside the bouding box of the person, the term $B$ is the person-bounded region, the term $C$ erases the pixel values within the bounding box  of the T-shirt, and the term $D$ is the newly introduced additive perturbation.   \textcolor{black}{In \eqref{eq: perturbation_plain},  the prior knowledge on $ M_{p,i} $ and $ M_{c,i} $ is acquired   by person detector and manual annotation, respectively.}
Without taking     into account physical transformations, 
% \textcolor{red}{[Xue: why by "without physical transformation" (4) reduces to the following? I see the difference is by excluding the person bounding box and only focusing on the cloth. I couldn't understand the difference as by without physical transformation.]}
Eq.\,\eqref{eq: perturbation_plain} simply reduces to the conventional formulation of adversarial example  
$(1-M_{c,i})\circ \mathbf x_i +  M_{c,i} \circ \boldsymbol{\delta}$.

% \textcolor{red}{[Xue: I feel the sentence is not completed. It should be continued with "to object detector such as in [ref][ref]"?}

Next, we   consider \textit{three main types} of physical transformations: a)   TPS  transformation   $t_{\mathrm{TPS}} \in \mathcal T_{\mathrm{TPS}}$   applying to the adversarial perturbation $\boldsymbol{\delta}$ for modeling the effect of cloth deformation, 
\textcolor{black}{b) physical color transformation $t_{\mathrm{color}}$ which converts digital colors to those printed and visualized in the physical world,}
and c) conventional physical transformation $ t \in \mathcal T$ applying to the   region within the person's bounding box, namely, $ (M_{p,i} \circ \mathbf x_i - M_{c,i} \circ \mathbf x_i + M_{c,i} \circ \boldsymbol{\delta})$. Here  $\mathcal T_{\mathrm{TPS}}$ denotes the set of possible non-rigid transformations,
\textcolor{black}{$t_{\mathrm{color}}$ is given by a  regression model  learnt from the color spectrum in the digital space to its printed counterpart,}
and 
$\mathcal T  $ denotes the set of commonly-used physical transformations, e.g., scaling, translation, rotation,   brightness, blurring and contrast. 
A modification of \eqref{eq: perturbation_plain} under different sources of transformations is then given by
{\small \begin{align}\label{eq: perturbation_robust}
    \mathbf x_i^\prime =& t_{\mathrm{env}} \left ( %\underbrace{ {(\mathbf 1 - M_{p,i}) \circ \mathbf x_i}}_{\text{A}}
    \text{A}
    + t \left ( \text{B}
    %\underbrace{ M_{p,i} \circ \mathbf x_i }_{\text{B}}
    - \text{C} %\underbrace{M_{c,i} \circ \mathbf x_i}_{\text{C}} 
    + t_{\mathrm{color}} ( M_{c,i} \circ t_{\mathrm{TPS}} (  \boldsymbol{\delta} + \mu \mathbf v ) )  \right ) \right )
    %\quad  t  \in \mathcal T, ~t_{\mathrm{TPS}} \in \mathcal T_{\mathrm{TPS}}, \mathbf v \sim \mathcal N(0,1),
\end{align}}%
for $t  \in \mathcal T$, $t_{\mathrm{TPS}} \in \mathcal T_{\mathrm{TPS}}$, and $\mathbf v \sim \mathcal N(0,1)$. In \eqref{eq: perturbation_robust},
\textcolor{black}{the terms A, B and C have been defined in \eqref{eq: perturbation_plain}, and $t_{\mathrm{env}}$ denotes a brightness transformation to model the environmental brightness condition. In \eqref{eq: perturbation_robust},
%$t(\cdot)$ depicts   the uncertainty of the physical environment,
%$t_{\mathrm{TPS}}(\cdot)$ is 
% $t_{\mathrm{TPS}}$ is learnt from the image pairs
% %obtained from  the TPS transformer \eqref{eq: TPSx} learnt under frame pairs 
% $\{ (M_{c,0} \circ \mathbf x_0 , M_{c,i} \circ \mathbf x_i) \}_{i=1}^$  for $i \in [M]$, and 
$\mu \mathbf v$ is an additive Gaussian   noise that allows the variation of pixel values, where $\mu $ is a given smoothing parameter and we set it as $0.03$ in our experiments such that the noise realization falls into the range $[-0.1, 0.1]$.  The randomized noise injection is also known as   Gaussian smoothing \cite{duchi2012randomized}, which   makes the final objective function  smoother and benefits the gradient computation during optimization.}
%\textcolor{Sijia_color}{[@Kaidi, Do we still have $\mathbf v$ in the new method?]}, 
%e.g., due to the mismatch between  the digital color and the printed color, 

%the conventional  $t_{\mathrm{person}}$ denotes a transformer applied to the image region characterized by a person's bounding box.
%in $\mathcal T$ except  $t_{\mathrm{env}}$.

% \textcolor{red}{[Xue: in (5), t() is applied including the mask, then why $t_{TPS}$() is applied excluding the mask?]}

% \begin{myremark}
\textcolor{black}{The prior work, e.g., \cite{sharif2016accessorize,evtimov2017robust}, established a  non-printability score (NPS) to measure the distance  between the designed perturbation vector and a library of printable colors acquired from the physical world. The commonly-used approach is to incorporate NPS into the attack loss through regularization. However, irt becomes non-trivial to find a proper regularization parameter, and the nonsmoothness of NPS  makes    optimization  for the adversarial T-shirt difficult.  To circumvent these challenges, we propose to model the color transformer $t_{\mathrm{color}}$ using 
a quadratic polynomial regression. The detailed color mapping is showed in Appendix\,\ref{app: color}.
}

With the aid of \eqref{eq: perturbation_robust}, the EoT formulation to fool a single object detector is cast as
\begin{align}\label{eq: attack_single}
    \begin{array}{ll}
\displaystyle \minimize_{\boldsymbol{\delta}}         &
\frac{1}{M}\sum_{i=1}^M \mathbb E_{t, t_{\mathrm{TPS}}, \mathbf v} \left [ f(\mathbf x_i^\prime )
\right ] + \lambda g(\boldsymbol{\delta}) 
%\\
    %\st      &  \boldsymbol{\delta} \in \mathcal C,
    \end{array}
\end{align}
where $f$ denotes an attack loss for misdetection,  
%we refer \cite{chen2018shapeshifter} and \cite{thys2019fooling} for our Faster R-CNN and YOLOv2 attack respectively.
$g$ is the total-variation norm that enhances perturbations' smoothness \cite{eykholt2018physical}, and  $\lambda > 0$ is a regularization parameter.
\textcolor{black}{We further elaborate on our attack loss $f$ in problem \eqref{eq: attack_single}. In YOLOv2, a probability score associated with a bounding box indicates  whether or not an object is present within this box. Thus, we specify the attack loss as the largest bounding-box probability  over   all   bounding boxes belonging to the   `person' class. For Faster R-CNN, we attack all bounding boxes towards the class `background'. 
The more detailed derivation on the attack loss is provided in Appendix\,\ref{app: loss}. Fig.\,\ref{fig: flowchart} presents an overview of our approach to generate adversarial T-shirts. 
% \textcolor{red}{[@Kaidi check. if you want you can remove the last sentence.]}
%this bound box include object.
}

\paragraph{Min-max optimization for fooling multiple object detectors.}
% It is  common  that the success rate of an attack will drop if the attack designed under one model is transformed to another model \textcolor{red}{[Refs]}.  
Unlike digital space, the transferability of adversarial  attacks  largely  drops in the physical environment, thus we  consider a  \textit{physical ensemble attack} against multiple object detectors.
It was recently shown in \cite{wang2019beyond} that the ensemble attack can be designed from the perspective of min-max optimization, and yields much higher worst-case attack success rate than the averaging strategy over multiple models. Given $N$ object detectors associated with attack loss functions $\{ f_i \}_{i=1}^N$, the {physical ensemble attack} is cast as
{\begin{align}\label{eq: attack_multiple}
    \begin{array}{ll}
\displaystyle \minimize_{\boldsymbol{\delta} \in \mathcal C}  \maximize_{\mathbf w \in \mathcal P}       &  \sum_{i=1}^N w_i \phi_i (\boldsymbol{\delta }) - \frac{\gamma}{2} \| \mathbf w - \mathbf 1/ N \|_2^2   + \lambda g(\boldsymbol{\delta}) , % \\
% \st          & \boldsymbol{\delta } \in \mathcal C,
    \end{array}
\end{align}}
where $\mathbf w$ are known as domain weights that adjust the importance of each object detector during the attack generation, 
$\mathcal P$ is a probabilistic simplex given by
$\mathcal P = \{\mathbf w | \mathbf 1^T \mathbf w = 1, \mathbf w \geq \mathbf 0 \}$, $\gamma > 0$ is a regularization parameter, 
and  $\phi_i (\boldsymbol{\delta}) \Def \frac{1}{M}\sum_{i=1}^M \mathbb E_{t \in \mathcal T, t_{\mathrm{TPS}} \in \mathcal T_{\mathrm{TPS}}} \left [ f(\mathbf x_i^\prime )
\right ]$ following   \eqref{eq: attack_single}. 
In \eqref{eq: attack_multiple}, if $\gamma = 0$, then the   adversarial perturbation $\boldsymbol{\delta}$ is designed over the \textit{maximum} attack loss (worst-case attack scenario)   since $ \maximize_{\mathbf w \in \mathcal P} \sum_{i=1}^N w_i \phi_i (\boldsymbol{\delta })  =\phi_{i^*} (\boldsymbol{\delta }) $, where $i^* = arg\,max_{i} \phi_i (\boldsymbol{\delta }) $ at a fixed $\boldsymbol{\delta}$. Moreover, if $\gamma \to \infty$, then the inner maximization of problem \eqref{eq: attack_multiple} implies $\mathbf w \to \mathbf 1/ N$, namely, an averaging scheme over $M$ attack losses. Thus, the regularization parameter $\gamma$ in \eqref{eq: attack_multiple} strikes a balance between the max-strategy and the average-strategy.

\section{Experiments}
% Object detectors: YOLOV2 and Faster R-CNN.
In this section, we demonstrate the effectiveness of our approach  (we call \textit{advT-TPS}) for design of the adversarial T-shirt 
by comparing it with  \textit{$2$} attack baseline methods,
a) adversarial patch to fool YOLOv2 proposed in \cite{thys2019fooling} and its printed version on a T-shirt (we call \textit{advPatch}\footnote{For fair comparison, we 
modify the perturbation size same as ours and execute the code provided in \cite{thys2019fooling} under our training dataset.}), and 
b) the variant of our approach in the absence of TPS transformation, namely, $\mathcal T_{\mathrm{TPS}} = \emptyset$ in \eqref{eq: perturbation_robust} (we call \textit{advT-Affine}). 
%During evaluation in the digital space, the simulated TPS is also applied. 
We examine the convergence behavior of   proposed algorithms as well as its  Attack Success Rate\footnote{ASR is given by the ratio of successfully attacked testing frames over the total number of  testing frames.} (ASR) in both digital and physical worlds. We 
clarify our algorithmic parameter setting   in Appendix\,\ref{app: hyper}. 

\textcolor{black}{Prior to detailed illustration, we briefly summarize the attack performance of our proposed adversarial T-shirt. 
When attacking YOLOv2, our method achieves 74\% ASR in the digital world and 57\% ASR in the physical world, where 
the latter is computed  by averaging successfully attacked video frames  over all 
different scenarios (i.e., indoor, outdoor and unforeseen scenarios) listed in Table\,2.
When attacking Faster R-CNN, our method achieves 61\% and 47\% ASR in the digital and the physical world, respectively. By contrast, the baseline advPatch only achieves around 25\% ASR in the best  case among all   digital and physical scenarios against either YOLOv2 or Faster R-CNN (e.g., 18\% against YOLOv2 in the physical case). 
% \QF{commented this out as it's against the anonymity policy.}
%We also remark that compared to our earlier version \cite{xu2019evading}, here the attack algorithm is  updated by incorporating color transformations, and the physical test videos are taken under multiple scenes. 
}

\subsection{Experimental Setup}

\paragraph{Data collection.}
\textcolor{black}{
  We collect two datasets  for learning and testing our proposed   attack algorithm in digital and physical worlds.
  %for generating adversarial T-shirts. 
The training dataset contains 
$40$ videos ($2003$ video frames) from $4$ different scenes: one outdoor and three indoor scenes.
each video takes $5$-$10$ seconds and was captured by  a moving  person   wearing a T-shirt with  printed checkerboard. 
% We use the first dataset to generate TPS transformations as well as learning 
The desired adversarial pattern is then learnt from the training dataset. 
The test dataset in the digital space contains $10$   videos captured under the same scenes as  the training dataset.  
This dataset is used to evaluate the attack  performance of the learnt adversarial pattern  in the digital world. In the physical world, we customize a T-shirt with the printed
 adversarial pattern learnt from our algorithm. Another $24$ test 
 videos (Section 4.3) are then collected at a different time capturing two or three persons (one of them wearing the adversarial T-shirt) walking a)  side  by  side  or b) at  different  distances. An  additional  control  experiment  in which     actors  wearing  adversarial T-shirts  walk  in  an  exaggerated  way  is  conducted to introduce large pose changes in the test data.
 In addition, we also test our adversarial T-shirt by unforeseen scenarios, where the test videos involve different locations and different persons which are never covered in the training dataset.
 All videos were taken using an iPhone X and resized  to 416 $\times$ 416.
 \textcolor{black}{In Table\,\ref{table:dataset} of the Appendix\,\ref{app:exp}, we summarize the collected dataset under all circumstances.}
% In the digital world, we use $30$ out of $40$ videos in the first dataset to learn the desired adversarial pattern, which is then tested under the remaining $10$ videos. 
}

% \begin{table}
%     \caption{Summarize of our collected dataset in each scenes.\textcolor{red}{[Move to supplement?]}}
%     \label{table:dataset}
%     \centering
%     \resizebox{0.9\linewidth}{!}{
%     \begin{tabular}{c|ccc|cc|c}
%         \toprule
%         \multirow{2}{*}{videos(frames)} &\multicolumn{3}{c|}{indoor} & \multicolumn{2}{c|}{outdoor} & overall\\
%         \cmidrule{2-6}
%                  &    office  & elevator  &  hallway  & street1 & street2 & \\
%         \midrule
%      single-person   & 4 (177) & 4 (135) & 4(230) & 4 (225) & 4 (240) & 20 (1007) \\
%      multi-persons  & 4 (162) & 4 (132) & 4(245) & 4 (230) & 4 (227)  & 20 (996)\\
%         \midrule
%      train           & 6 (245) & 6 (180) & 6(335) & 6 (344) & 6 (365) & 30 (1469) \\
%      test (digital)   & 2 (94) & 2 (87) & 2(140) & 2 (111) & 2 (102) & 10 (534) \\
%      & 
%      test (physical)   & XXX (20) & XXX (16) & XXX(30) & 2 (30) & XXXX (30) & XXX (504) \\
%         \bottomrule
%     \end{tabular}
%     }
% \end{table}

\paragraph{Object detectors.}
We use two state-of-the-art object detectors: Faster R-CNN~\cite{ren2015faster} and YOLOv2~\cite{redmon2017yolo9000} to evaluate our method. These two object detectors are both pre-trained on COCO dataset~\cite{lin2014microsoft} which contains 80 classes including `person'.  The detection minimum threshold are set as  0.7 for both Faster R-CNN and YOLOv2 by default. The sensitivity analysis of this threshold is performed in Fig.\,\ref{fig: threshold} Appendix\,\ref{app: hyper}.

\subsection{\textcolor{black}{Adversarial T-shirt in the digital world}}\label{exp: digital}

% \textcolor{Sijia_color}{[Needs a better organization on experiments. Highlights: a) Convergence of algorithm to generate  adversarial T-shirt attack. Loss versus iterations, and visualization, etc. b) Attack performance in the digital world. Giant table for ASR to compare our method with baselines, including
% b-1) The need of TPS, b-2) transferability and min-max ensemble attack, b-3) other attacks. And associated figures corresponding to b-1), b-2), b-3). c) Attack performance in the physical world (T-shirt attached with adversarial pattern printed on paper). Choose single-detector attack, and compared to baseline. Choose multiple-detector ensemble attack, and compared to baseline. d) authentic adversarial T-shirt with printed adversarial pattern.]}

\begin{figure}[htb]
     \begin{center}
     \begin{tabular}{cc}
    \includegraphics[width = 0.43\textwidth]{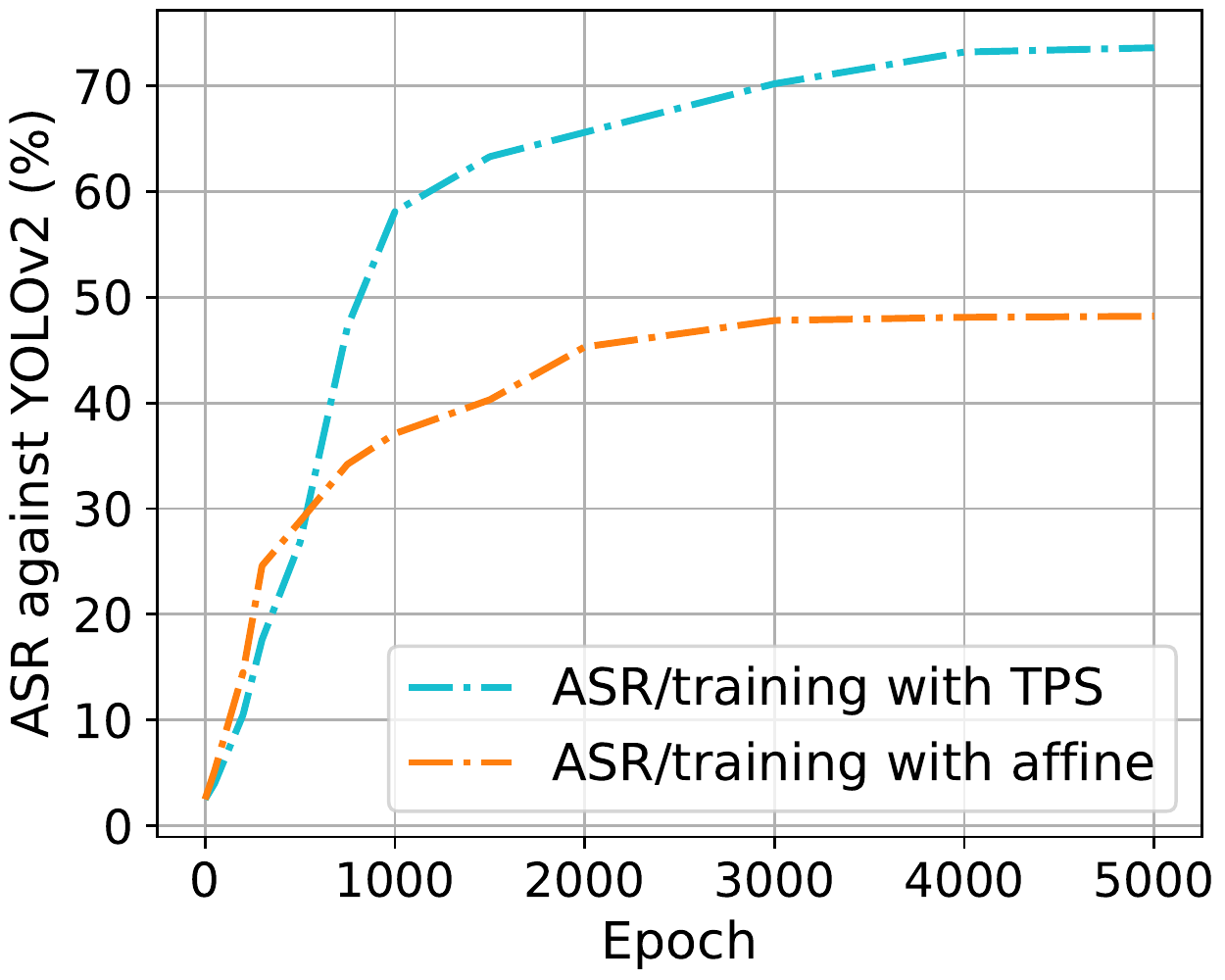}
    \hspace{4mm}\includegraphics[width = 0.43\textwidth]{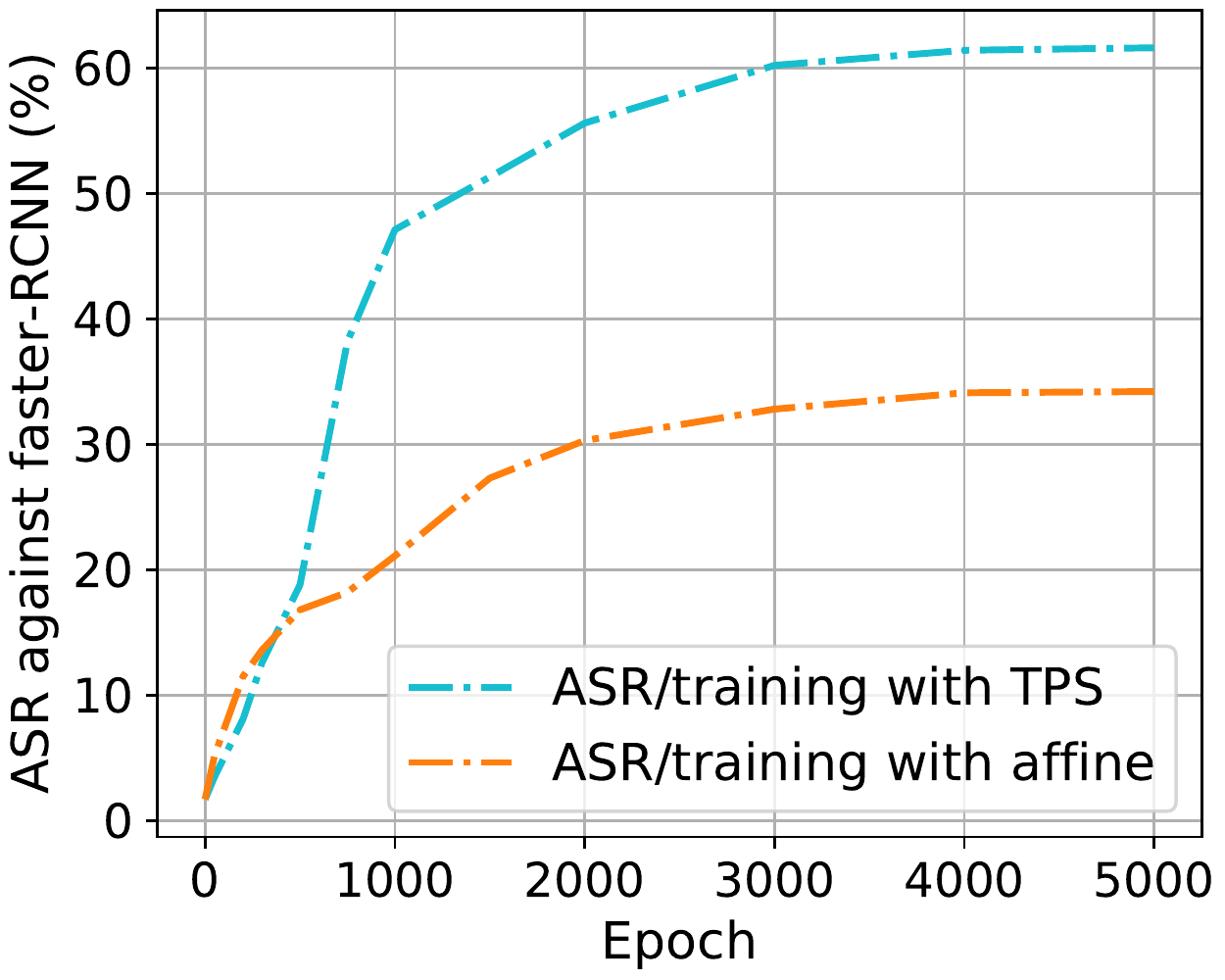}
      \end{tabular}
      \vspace{-2mm}
      \caption{{ASR v.s. epoch numbers against YOLOv2 (left) and Faster R-CNN (right).
      }}
      \label{tab: convergence}
      \end{center}
      \end{figure}

\paragraph{Convergence performance of our proposed attack algorithm.}
%[Convergence of algorithm to generate  adversarial T-shirt attack. Loss versus iterations, and visualization, etc]
In Fig.~\ref{tab: convergence}, we show ASR against the epoch number used by our proposed algorithm to solve problem \eqref{eq: attack_single}.  Here the success of our attack at one testing frame is required to meet two conditions, a)  misdetection of the person who wears the adversarial T-shirt, and b)   successful detection of the person whom dresses a normal cloth. 
% the person with the digital adversarial T-shirt is not detected by object detector at all and the control person (with normal clothes) is detected at the meantime.
As we can see, the proposed attack method covnerges well for attacking both YOLOv2 and Faster R-CNN.
We also note that attacking Faster R-CNN is more difficult than attacking YOLOv2. Furthermore, if TPS is not applied during training, then ASR drops around $30\%$ compared to our approach by leveraging TPS.

% \vspace*{-4mm}

\paragraph{ASR of adversarial T-shirts in various attack settings.}
%[Giant table for ASR to compare our method with baselines, including b-1) The need of TPS, b-2) transferability and min-max ensemble attack.  And associated figures corresponding to b-1), b-2)]

We perform a more comprehensive evaluation on our methods by digital simulation.  Table~\ref{table:TPS_models}   compares the ASR of   adversarial T-shirts generated w/ or w/o TPS transformation in 4 attack settings: 
a) \textit{single-detector attack} referring to adversarial T-shirts designed and evaluated using the same object detector, b) \textit{transfer single-detector attack} referring to adversarial T-shirts designed and evaluated using different object detectors, c) \textit{ensemble attack (average)} given by  \eqref{eq: attack_multiple} but  using the average of attack losses of individual models, and d) 
 \textit{ensemble attack (min-max)} given by \eqref{eq: attack_multiple}.
%when generate perturbation with/without TPS transformation.
% As we can see generate perturbation by applying TPS transformation only drop slightly (1\%) when testing without TPS transformation. However, when we simulate applying TPS transformation in testing stage, the ASR of perturbation generate without considering TPS transformation drop from 0.83 to 0.52 when attacking YOLOv2, compare with generating with TPS transformation, the ASR hold well. 
% It demonstrate that TPS transformation is significant for an adversarial T-shirt at least in digital world.
% In order to make the comparison fair enough, we also test the random noise perturbation which will cause 12\% and 15\% ASR for Faster R-CNN and YOLOv2 respectively.
As we can see,  it is crucial to incorporate TPS transformation in the design of adversarial T-shirts: without TPS, the ASR drops from 61\% to 34\% when attacking faster R-CNN and drops
from 74\% to 48\% when attacking YOLOv2 in the single-detector attack setting.
We also note that the transferability of single-detector attack is   weak in all settings. 
And faster R-CNN is consistently more robust than YOLOv2, similar to the results in Fig.\,\ref{tab: convergence}. Compared to our approach and  \textit{advT-Affine},  the baseline method \baseline~yields the worst ASR when attacking a single detector. 
Furthermore, we evaluate the  effectiveness of the proposed {min-max} ensemble attack \eqref{eq: attack_multiple}.  
% Please note that, if we simply transfer attack the two detectors with the perturbation generated by the other one, in Table~\ref{table:TPS_models}, the ASR are only around 10\% in all cases which similar with the results of random noise. So the attack transferability or Faster R-CNN and YOLOv2 are weak.  
% For comparison, we present ASR of the convectional ensemble attack by \textit{averaging} the attack losses  over multiple models with our \textit{min-max} way.
As we can see,   when attacking faster R-CNN, 
the min-max ensemble attack significantly outperforms its counterpart using the averaging strategy, leading to $15\%$ improvement in ASR.
This improvement is at the cost of  $7\%$ degradation  when attacking YOLOv2.

\begin{table}[htb]
\vspace{-4mm}
    \centering
            \caption{The ASR ($\%$) of adversarial T-shirts generated from our approach, \affine and the baseline \baseline~ under digital-world.}
            \adjustbox{max width=1\textwidth}{
    \begin{tabular}{c|c|c|c|c|c}
        \toprule[1pt]
         method & model  & target & transfer  & ensemble(average) & ensemble(min-max)  \\
        \midrule[1pt]
        advPatch\cite{thys2019fooling}       &        & 22\% & 10\%  & N/A & N/A  \\
        advT-Affine & Faster R-CNN   & 34\% & 11\%  & 16\%  & 32\%  \\
        advT-TPS(ours)  &            & \textbf{61\%} & 10\%  & \textbf{32\%}  & \textbf{47\%}  \\
        \midrule[1pt]
        advPatch\cite{thys2019fooling}       &        & 24\% & 10\%  & N/A  & N/A  \\
        advT-Affine & YOLOv2         & 48\% & 13\%  & 31\%  & 27\%  \\
        advT-TPS(ours)  &            & \textbf{74\%} & 13\%  & \textbf{60\%}  & \textbf{53\%}  \\
        \bottomrule[1pt]
    \end{tabular}
        }
    \label{table:TPS_models}
    \vspace{-4mm}
\end{table}

\subsection{Adversarial T-shirt in the physical world}\label{exp:physical}
% [ Choose single-detector attack, and compared to baseline ]

%\textcolor{Sijia_color}{[Rewrite this. Very very difficult to follow. The result are also not consistent. First, please write physical setting. E.g., Generating adversarial pattern using our method to attack faster R-CNN...... Second, mention baseline. Third, have a table on ASR. Forth, talk about visualization result.]}

We next evaluate our method in the physical world. First, we generate an adversarial pattern by solving problem \eqref{eq: attack_single} against YOLOv2 and Faster R-CNN, following Section~\ref{exp: digital}. 
We then print the adversarial pattern on a white T-shirt, leading to the adversarial T-shirt. For fair comparison, we also print  adversarial patterns  generated by the \baseline~\cite{thys2019fooling}  and \affine in Section\,\ref{exp: digital} on white T-shirts of the same style. 
% We craft adversarial T-shirt by first printing perturbation and then paste it on a white T-shirt.  
% \textcolor{blue}{We didn't mention how other T-shirts are generated and printed }
% \QF{Kaidi, please check this sentence. it's different from what we wrote in the experimental setup.}
It is worth noting that different from evaluation by  taking static photos of physical adversarial examples, our evaluation is conducted at a more practical and   challenging setting. That is because   we  record  videos to track a moving person wearing  adversarial T-shirts, which could encounter multiple environment effects
% at the testing phase, we  record  videos to track a moving person wearing  adversarial T-shirts in various scenarios. By contrast with taking static photos, our evaluation over an entire video takes into account multiple environment effects 
such as distance, deformation of the T-shirt, poses and angles of the moving person.
%is captured by a iPhone 5 at the similar environment we used in training dataset.
%In experiment, 

In Table\,\ref{table:physical},  
we compare our method with   \baseline  and \textit{advT-Affine} under $3$ specified scenarios, including the indoor, outdoor, and unforeseen scenarios\footnote{Unforeseen scenarios refer to test videos  that  involve different locations and   actors, never seen in the training dataset.}, together with the overall case of all scenarios.
 We observe that our method achieves 64\% ASR (against YOLOv2), which is much higher than \affine (39\%)  and \baseline (19\%) in the indoor scenario. 
 Compared to the indoor scenario, evading person detectors in the outdoor scenario becomes more challenging. The ASR of our approach reduces to 47\% but outperforms \affine (36\%) and \baseline (17\%).
%  In most cases, our methods yield better performance in indoor scenario and slightly worse performance in outdoor scenario. 
 This is not surprising since the   outdoor scenario suffers   more environmental variations such as lighting change.
 Even considering the unforeseen scenario, we find that
  our adversarial T-shirt is robust to the change of person and location, leading to 48\% ASR against Faster R-CNN and 59\% ASR against YOLOv2. Compared to the digital results, 
the ASR of our adversarial T-shirt drops around $10\%$ in all tested physical-world scenarios; see specific video  frames in Fig.\,\ref{fig: physical_examples}.
% Note that the attacking loss of Faster R-CNN for baseline method is not provided in their work, so we conduct a transfer attack which the perturbation generated by attacking YOLOv2.

%  \QF{the last column is the average ASR on all the scenes or an overall ASR? An overall ASR is better. In the text, this is mentioned as average ASR. Please double check}

\begin{table}[htb]
\vspace{-4mm}
    \centering
            \caption{The ASR ($\%$) of adversarial T-shirts generated from our approach, \affine and \baseline under different physical-world scenes.}
            \adjustbox{max width=1\textwidth}{
    \begin{tabular}{c|c|c|c|c|c}
        \toprule[1pt]
         method & model  & indoor & outdoor & new scenes & average ASR  \\
        \midrule[1pt]
        advPatch\cite{thys2019fooling}       &        & 15\% & 16\%  & 12\%  & 14\%  \\
        advT-Affine & Faster R-CNN   & 27\% & 25\%  & 25\%  & 26\%  \\
        advT-TPS(ours)  &            & \textbf{50}\% & \textbf{42\%}  & \textbf{48\%}  & \textbf{47\%}  \\
        \midrule[1pt]
        advPatch\cite{thys2019fooling}       &        & 19\% & 17\%  & 17\%  & 18\%  \\
        advT-Affine & YOLOv2         & 39\% & 36\%  & 34\%  & 37\%  \\
        advT-TPS(ours)  &            & \textbf{64\%} & \textbf{47\%}  & \textbf{59\%}  & \textbf{57\%}  \\
        \bottomrule[1pt]
    \end{tabular}
    }
    \label{table:physical}
    \vspace{-4mm}
\end{table}

\subsection{Ablation Study}\label{sec:abalation}
% \QF{start here ====}

In this section, we conduct more experiments for better understanding the robustness of our adversarial T-shirt against various conditions including   angles and  distances to camera, camera view, person's pose, and complex scenes that include crowd and occlusion.
Since the baseline method (\baseline) performs poorly in most of these scenarios, we focus on evaluating our method (\TPS) against \affine using YOLOv2. 
\textcolor{black}{We refer readers to Appendix\,\ref{app:exp_setup} for details on the setup of our ablation study.}

% \textcolor{red}{[Move this to appendix.]}
% In order to analyse the sensitivity of each controllable variable, we recollected new data for all the experiments below. Specifically, we used five people (two females and three males) for this study and none of them appeared in the training and physical-world test data. For each condition aforementioned, we recorded one or multiple videos for each person. As opposed to the previous physical-world evaluation, these experiments (except the distance one) have all the actors stay at a fixed location, usually 1-3m away from the camera. 

%\textcolor{red}{[SL: Distance and view can be combined to one paragraph.]}

\begin{figure}[h!]
     \begin{center}
     \begin{tabular}{ c c}
    \includegraphics[width = 0.43\textwidth]{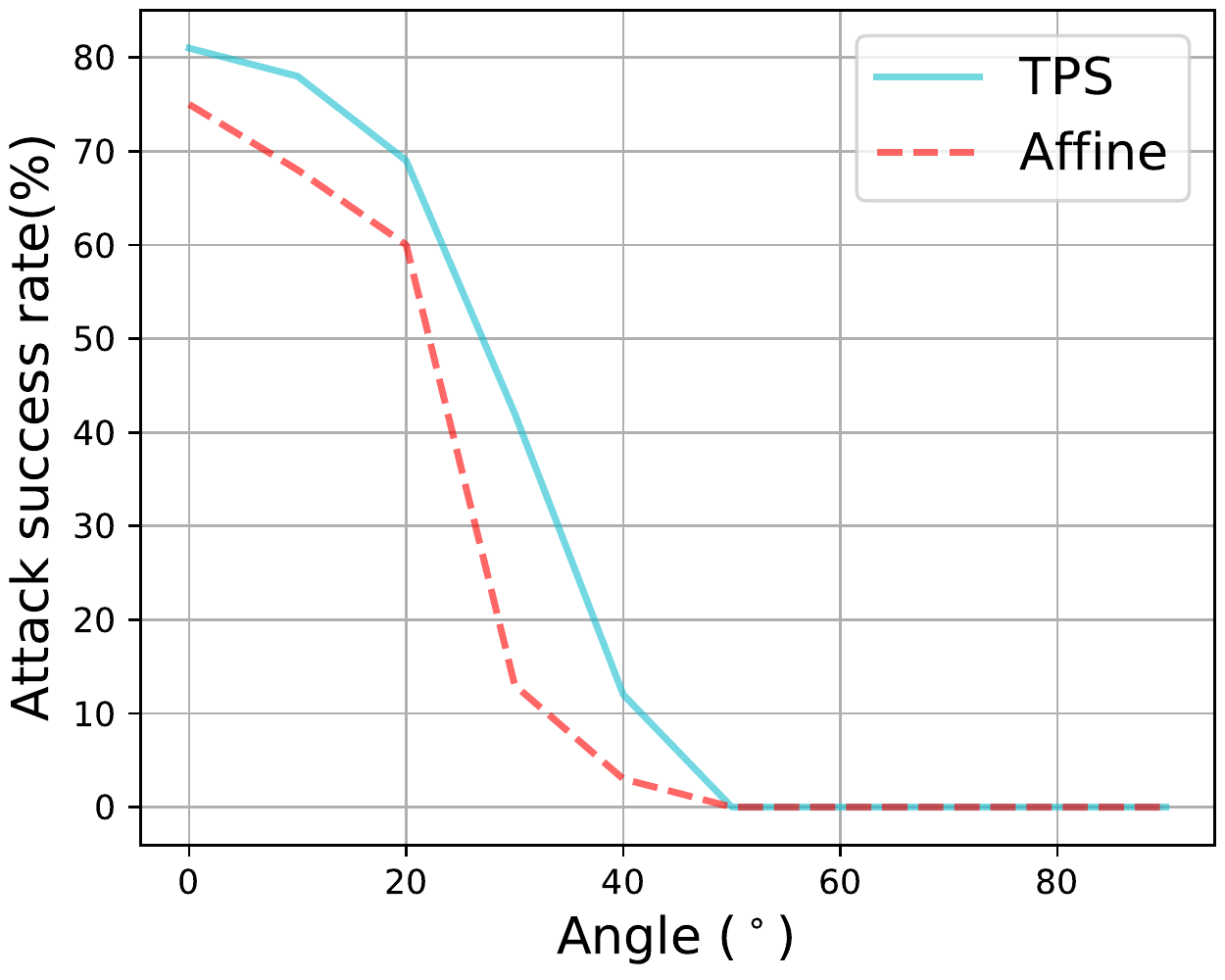}
    \hspace{4mm}\includegraphics[width = 0.43\textwidth]{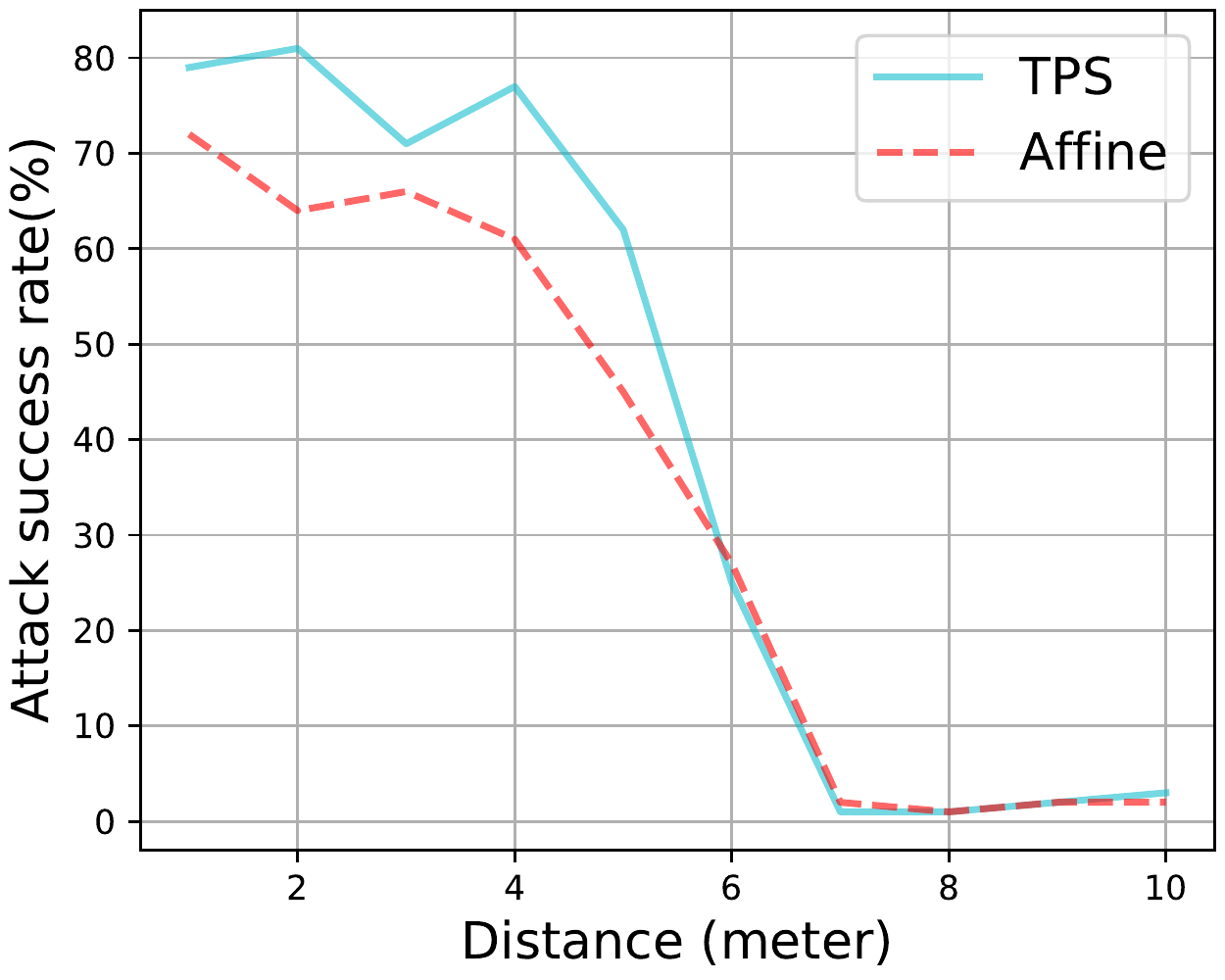}
      \end{tabular}
      \caption{{Average ASR v.s. different angles (left) and distance (right).
      }}
      \label{fig:distance_angle}
      \end{center}
      \vspace{-6mm}
      \end{figure}

\paragraph{Angles and distances to camera.} 
 In Fig.\,\ref{fig:distance_angle}, we present ASRs of \TPS and \affine when the actor whom wears the adversarial T-shit at different angles and distances to the camera. As we can see, \TPS   works well within the  angle  $20^\circ$ and the distance $4$m. And \TPS consistently outperforms  \affine. We also note that ASR drops significantly at the  angle  $30^\circ$ since it induces  occlusion of the adversarial pattern. Further, if the distance is greater than $7$m, the pattern cannot clearly be seen  from the camera. 
% shows that the our adversarial T-shirt works well within a camera angle of $20^\circ$. The efficacy is reduced by half at $30^\circ$ and almost lost after $40^\circ$. Note that the adversarial pattern undergoes occlusion already at $40^\circ$\QF{Kaidi, please check the accuracy of this info.}, thus resulting in a significant drop of ASR. We expect to improve the robustness of the T-shirt against camera view change by enriching the training data with more camera view changes.

    %   \vspace*{-4mm}

% \textbf{Camera distance}. We test how the distance of a camera from a person affects the model performance when the person faces the camera upright. The test distance ranges from 1m to 10m. As can be seen from Fig.\,\ref{fig:distance_angle}, the adversarial T-shirt stays effective when the person is within 4m to the camera. Beyond that distance the ASR drops dramatically till 8m where the T-shirt completely fails to fool the detector. \QF{Sijia, please check this comment. Feel free to rewrite it.} We comment that far-view cameras are still a challenge our current approach due to low an image resolution. However, a distance of 4$\sim$8m are an effective range for attacking mid-view or close-view cameras, which are often used for indoor scenarios such as offices and building entrances/exits.

% \textbf{Camera view}. Given that security cameras are usually hidden in unnoticeable places, it is important to ensure that the adversarial T-shirt is tolerant to some camera view changes to some degree.  Fig.\,\ref{fig:distance_angle} shows that the our adversarial T-shirt works well within a camera angle of $20^\circ$. The efficacy is reduced by half at $30^\circ$ and almost lost after $40^\circ$. Note that the adversarial pattern undergoes large deformation already at $40^\circ$, thus resulting in a significant drop of ASR. We expect to improve the robustness of the T-shirt against camera view change by enriching the training data with more camera view changes.

\begin{figure*}[htb]
   \centering
   \adjustbox{max width=1\textwidth}{

\begin{tabular}{p{0.1in}p{0.53in}p{0.53in}p{0.53in}|p{0.53in}p{0.53in}p{0.53in}|p{0.53in}p{0.53in}}

& \multicolumn{3}{c|}{ \footnotesize crouching} & \multicolumn{3}{c|}{ \footnotesize sitting} &\multicolumn{2}{c}{ \footnotesize running}
\\
\rotatebox{90}{\parbox{0.9in}{\centering \footnotesize \affine}}&
\includegraphics[width=0.54in]{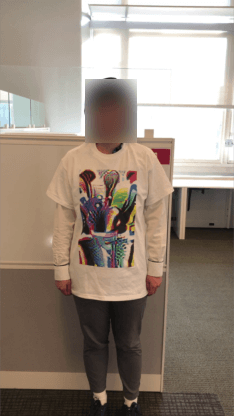}& 
\includegraphics[width=0.54in]{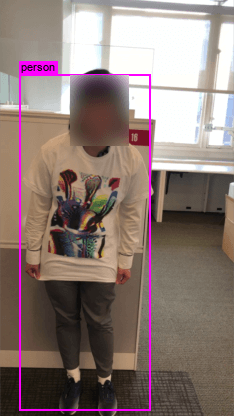}& 
\includegraphics[width=0.54in]{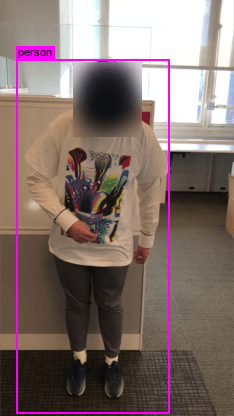}& 
\includegraphics[width=0.54in]{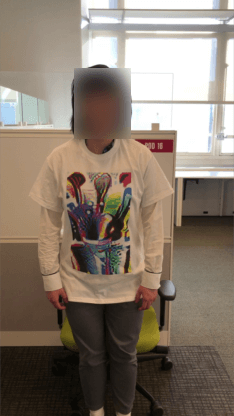}& 
\includegraphics[width=0.54in]{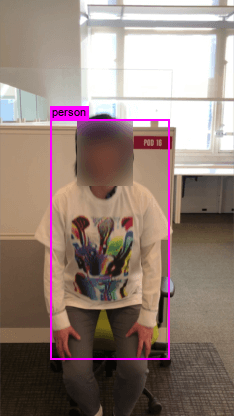}& 
\includegraphics[width=0.54in]{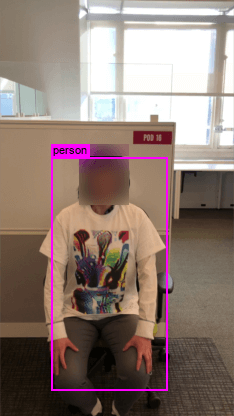}& 
\includegraphics[width=0.54in]{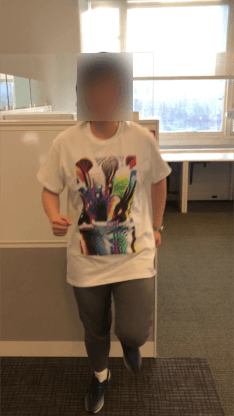}& 
\includegraphics[width=0.54in]{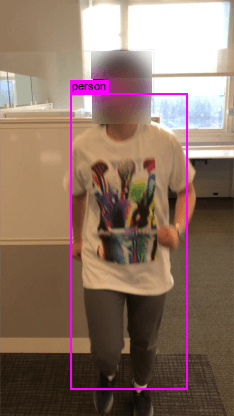}
\\
\rotatebox{90}{\parbox{0.9in}{\centering \footnotesize \TPS}}&
\includegraphics[width=0.54in]{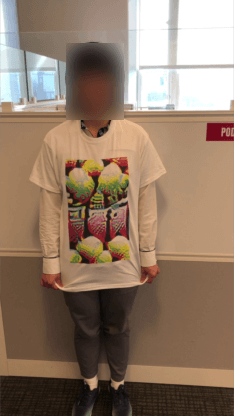}& 
\includegraphics[width=0.54in]{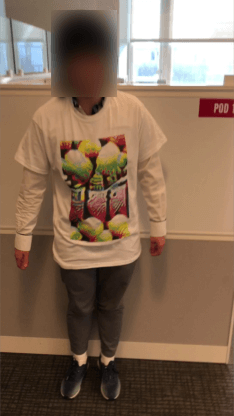}& 
\includegraphics[width=0.54in]{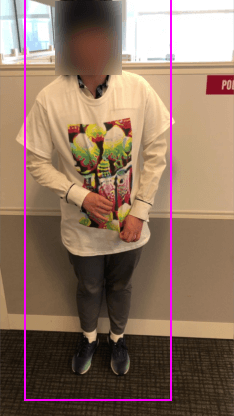}& 
\includegraphics[width=0.54in]{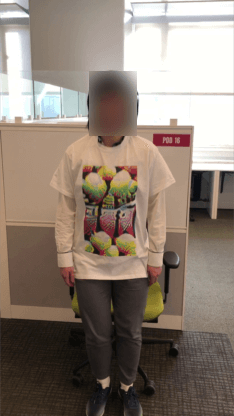}& 
\includegraphics[width=0.54in]{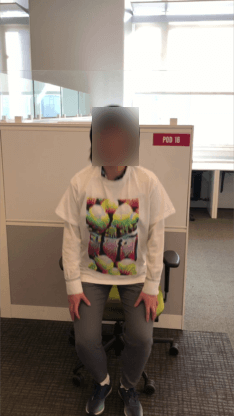}& 
\includegraphics[width=0.54in]{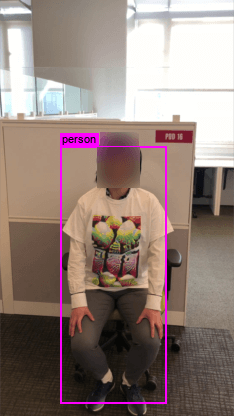}& 
\includegraphics[width=0.54in]{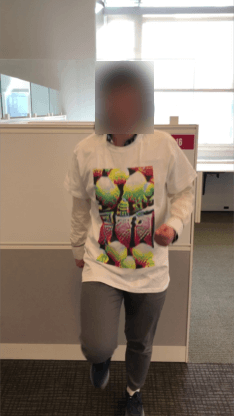}& 
\includegraphics[width=0.54in]{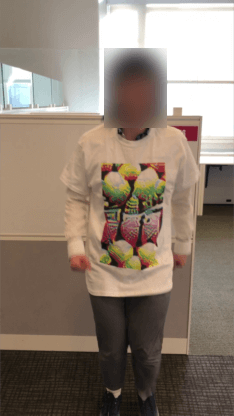}
\\
\end{tabular}
}
% \vspace{-2mm}
\caption{\footnotesize{
Some video frames of person who wears adversarial T-shirt generated by \affine (first row) and \TPS (second row) with different poses.
}}
    \label{fig: pose}
    %\vspace*{-4mm}
\end{figure*}

\paragraph{Human Pose.} In  Table\,\ref{table:pose} (left), we evaluate
  the effect of pose change on \TPS, where videos are taken for an actor  with  some distinct postures including  crouching,  siting  and  running  in place; see Fig.\,\ref{fig: pose} for specific examples. To  alleviate  other  latent  effects,  the camera was made to look straight at the person at a fixed distance of about $1\sim2$m away from the person. 
  As we can see, \TPS consistently outperforms \affine. However, it is worth noting that the sitting posture remains challenging for both methods as the larger occlusion is the worse ASR is. To delve into this problem, {Fig.\,\ref{fig: occlusion}}   presents how well our adversarial T-shirt can handle occlusion by partially covering the T-shirt by hand.
 Not surprisingly, both \affine and \TPS may fail when occlusion becomes quite large.   Thus,  occlusion is still an interesting problem for physical adversaries. %Nevertheless, we test how well our adversarial T-shirt can handle occlusion by partially covering the T-shirt by hand.  We leave this challenging problem for future research.
%  \textcolor{red}{We refer readers to Table\,\ref{table:pose}(left) for more comprehensive  statistics on ASR against person's pose change.}
%As shown in  Table 5  (left), our TPS-based approach significantly outperforms the affine-based approach on the crouching and running poses, clearly demonstrating the robustness of TPS against pose changes over the affine transformation. It is also observed that the sitting posture remains challenging for our approach to handle, largely because that sitting likely results in large wrinkles at the bottom of the T-shirt.

\paragraph{Complex scenes.} In Table\,\ref{table:pose} (right), we test our adversarial T-shirt in several complex scenes with cluttered backgrounds, including a) an office with multiple objects and people moving around; b) a parking lot with vehicles and pedestrians; and c) a crossroad with busy traffic and crowd. We observe that compared to \affine, \TPS is reasonably effective in complex scenes without suffering a significant loss of ASR.
Compared to the other factors such as camera angle and occlusion, cluttered background and even crowd are probably the least of a concern for our approach. This is explainable, as our approach works on object proposals directly to suppress the classifier. 
% As a result, the capability of the detector for detecting objects in complex scenes remains unattacked. 
% For the same reason, \affine demonstrates ...

%not surprising as detectors 
%our approach remains reasonably effective even these scenes are much more c, in this experiment, person who wear adversarial T-shirt need to keep static so that actually both two methods yield acceptable performance. Moreover, if we compare the ASR in Table\,\ref{table:pose}, the ASR are much better when we only change environment, it demonstrates that the complexity of background has minor effect on ASR compare with the pose of person. The examples of our complex scenes are shown in Fig.\,\ref{tab: complex_scene}. \textbf{Other conditions.} In the physical world, some environment variables such as brightness and crowd can be covered in complex scenes. Especially in outside environment, the brightness are changed obviously (see Fig.\,\ref{tab: complex_scene}) caused by the cloudy and time. 

%the  average ASR of TPS are much better than Affine, especially in crouching and running cases. Because affine transformation cannot perform deformation like crouching so that slightly crunching may lead attack useless. Some examples are performed in Fig.\,\ref{fig: pose}. We refer readers to see videos in supplementary files. For the siting condition, because some evident wrinkle happened in lower part of the T-shirt,

\begin{table}[htb]
 \centering
  \caption{The ASR ($\%$) of adversarial T-shirts generated from our approach, \affine and \baseline under different physical-world scenarios. 
 }
\adjustbox{max width=1\textwidth}{

\begin{tabular}{|c|c|c|c|}
\toprule[1pt]
\backslashbox{Method}{Pose}  & crouching  &  siting & running              \\ 
\midrule[1pt]
\affine   & 27\%    &  26\%     & 52\% \\
\TPS      & \textbf{53\%}      & \textbf{32\%}  &\textbf{63\%}\\
\bottomrule[1pt]
\end{tabular}
\,
% \end{minipage}%
%  \begin{minipage}{.5\linewidth}
\begin{tabular}{|c|c|c|c|}
\toprule[1pt]
\backslashbox{Method}{Scenario}  & office  &   parking lot   &   crossroad    \\ 
\midrule[1pt]
\affine   & 69\%    &  53\%     &  51\%   \\
\TPS      & \textbf{73\%}  &\textbf{65\%}  &\textbf{54\%}\\
\bottomrule[1pt]
\end{tabular}
}
 \label{table:pose}
\end{table} 

% To test the complexity of environment, we take videos in two scenes 1) office with multiple objects and other moving or fixed persons; 2) parking lot with multiple cars and persons. The results can be found in Table\,\ref{table:pose} (right). Consider to mitigate the effect of other controllable variables, in this experiment, person who wear adversarial T-shirt need to keep static so that actually both two methods yield acceptable performance. Moreover, if we compare the ASR in Table\,\ref{table:pose}, the ASR are much better when we only change environment, it demonstrates that the complexity of background has minor effect on ASR compare with the pose of person. The examples of our complex scenes are shown in Fig.\,\ref{tab: complex_scene}.

\begin{figure}[htb]
     \begin{center}
     \begin{tabular}{ c c c}
    \includegraphics[width = 0.2\textwidth]{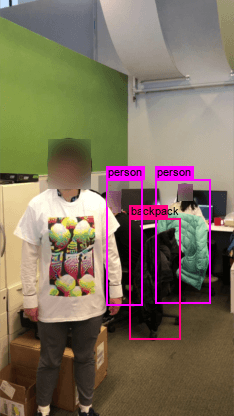}
    \includegraphics[width = 0.2\textwidth]{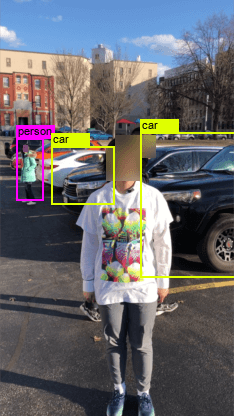}
    \includegraphics[width = 0.2\textwidth]{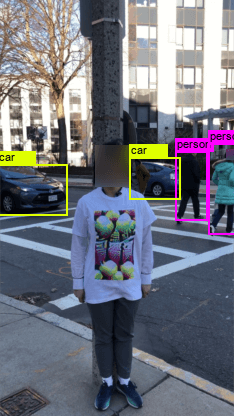}

      \end{tabular}
      \caption{\footnotesize{The person who wear our adversarial T-shirt generate by TPS in three complex scenes: office, parking lot and crossroad.
      }}
      \label{tab: complex_scene}
      \end{center}
      \end{figure}

\section{Conclusion}

% \textcolor{Sijia_color}{@Kaidi}
In this paper, we propose \textit{Adversarial T-shirt}, the first successful  adversarial wearable to evade detection of moving persons. Since T-shirt is a non-rigid object, its deformation induced by a person's pose change  is taken into account when generating adversarial perturbations. 
We also propose a   min-max ensemble  attack algorithm to fool multiple object detectors simultaneously.
We show  that  our attack  against YOLOv2 can achieve 74\% and {57\%}  attack success rate in the digital and physical world, respectively.  By contrast, the \baseline method can only achieve 24\% and {18\%}  ASR. 
Based on our studies, we hope to provide 
 some implications on how the adversarial perturbations can be implemented in physical worlds.
%  with human clothing, accessories,  paint on face, and other wearables, and we also aim to establish a general framework for evaluating the robustness of real-time machine learning systems deployed in physical worlds.

% \clearpage
\bibliographystyle{splncs04}
\bibliography{egbib}

\input{appendix}
\end{document}

%% file: math_commands.tex
\usepackage{amsmath,amsfonts,bm}

\def\eqref#1{(\ref{#1})}
\def\1{\bm{1}}

\DeclareMathAlphabet{\mathsfit}{\encodingdefault}{\sfdefault}{m}{sl}
\SetMathAlphabet{\mathsfit}{bold}{\encodingdefault}{\sfdefault}{bx}{n}

%% file: appendix.tex
\def\httilde{\mbox{\tt\raisebox{-.5ex}{\symbol{126}}}}
\setcounter{section}{0}
\setcounter{figure}{0}
\renewcommand{\thefigure}{A\arabic{figure}}
\makeatother
\setcounter{table}{0}
\renewcommand{\thetable}{A\arabic{table}}

% \begin{document}
\appendix

\clearpage
\newpage
\section*{Appendix}

\textcolor{black}{In the supplement, we provide details on  the thin plate spline (TPS) transformation, the formulation of attack loss, the setting of algorithmic parameters, and the additional   experiments of the   adversarial T-shirt in the physical world. 
% Prior to detailed illustration, we briefly summarize the attack performance of our proposed adversarial T-shirt. 
% When attacking YOLOv2, our method achieves 74\% attack success rate (ASR) in the digital world and 57\% ASR in the physical world, where 
% the latter is computed  by averaging successfully attacked video frames  over all 
% different scenarios (i.e., indoor, outdoor and unforeseen scenarios) listed in Table\,2.
% When attacking Faster R-CNN, our method achieves 61\% and 47\% ASR in the digital and the physical world, respectively. By contrast, the baseline method (namely, the T-shirt printed using the physical adversarial patch~\cite{thys2019fooling} designed for evading   person detectors) only achieves around 25\% ASR in the best  case under both digital and physical scenarios.
} 

% our attack  against YOLOv2 can achieve 74\% and 57\%  attack success rate in the digital and physical world, respectively.  By contrast, the baseline method can only achieve 24\% and \textcolor{Sijia_color}{18\%}  ASR. 

\section{How to construct TPS transformation?}\label{app: TPS}

\begin{figure}[h!]
     \begin{center}
     \begin{tabular}{ cc}
     \hspace{5mm}   frame 1 in Fig.2a & frame 2 in Fig.2b\\
    \multicolumn{2}{c}{\includegraphics[width = 0.4\textwidth]{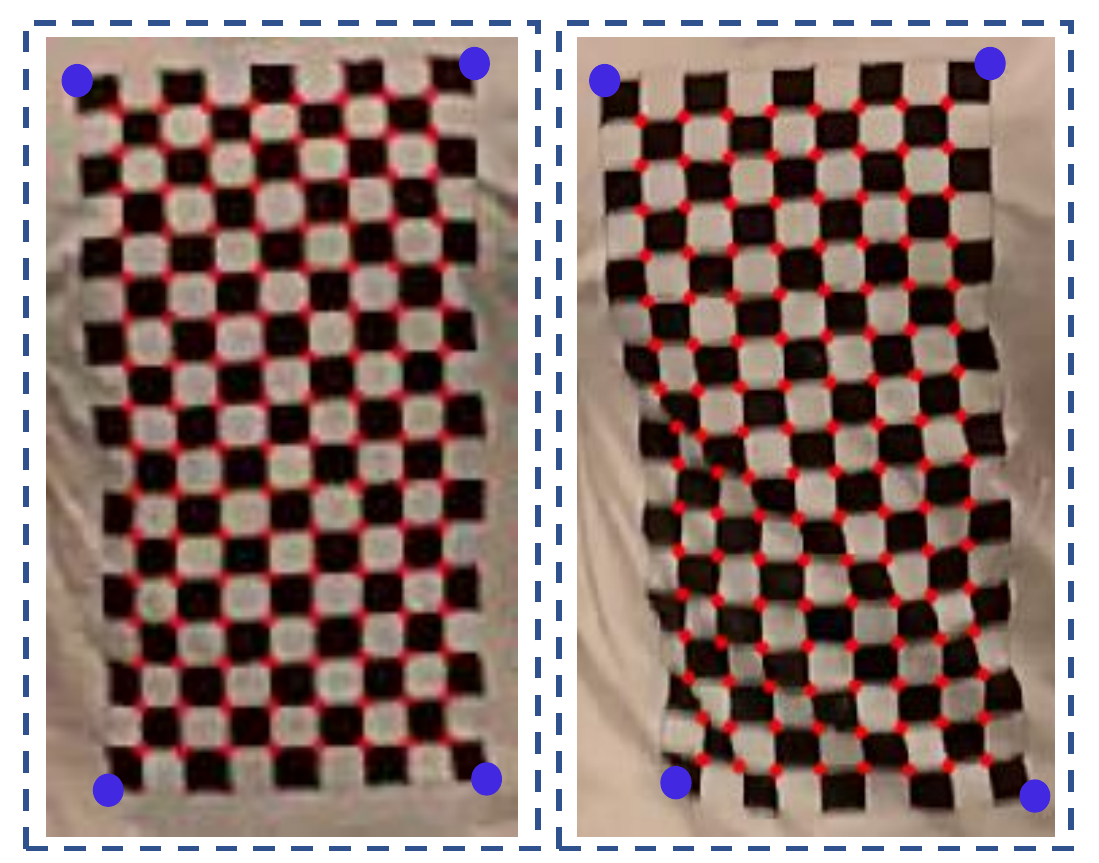}}
      \end{tabular}
      \vspace{2mm}
      \caption{\textcolor{black}{Four manually annotated corner points (blue) used to generate the bounding box of cloth region at frame $i$, namely, $M_{c,i}$. And  $8 \times 16$ anchor points (red) on the checkerboard used to generate TPS transformation $t_{\mathrm{TPS}}$ between two video frames.
      }}
      \label{fig: anchors}
      \end{center}
      \end{figure}

 We first manually annotate four corner points (see blue markers in Figure~\ref{fig: anchors})   to conduct a
perspective transformation between two frames at  different time instants.
This perspective transformation is used to align the coordinate system of anchor points used for TPS transformation between two frames. 

      \textcolor{black}{Ideally, the checkerboard detection tool~\cite{geiger2012automatic,zhang2000flexible} always outputs a grid of corner points detected. In most cases, it can locate all the $8\times16$ points on the checkerboard perfectly, so no additional effort is needed to establish the point correspondences between two images. In the case when there are corner points missing in the detection, we use the following method to match two images.
      %(we need a non-perfect images here to illustrate the case where there are corner points missing in the detection). 
      }
\textcolor{black}{
% coordinate system alignment followed
% by point aliment, where the former refers to conducting a
% perspective transformation from one frame to the other
% . Then we normalize the two checkerboard area into the same coordinate system by the two pairs of outer anchor points.
We perform a point matching procedure (see Algorithm~\ref{alg: TPS}) to align the anchor points (see red markers in Figure~\ref{fig: anchors}) detected by the checkerboard detection tool.
% Ideally, if the checkerboard detect tool runs perfectly, we can find all $8 \times 16$ anchor points from all target images. However, due to the deformation of T-shirts, the detected   anchor points might be  incomplete  at different video frames.  Thus,
The data matching procedure selects the set of matched anchor points  used for constructing TPS transformation.} 

% After that, we collect $8 \times 16$ anchor points in all images from the checkerboard by a checkerboard detect tool~\cite{geiger2012automatic} as shown in Figure~\ref{fig: anchors} in red. Ideally, if the checkerboard detect tool running perfectly we can achieve $8 \times 16$ anchor points from all target images. However, because some images have extreme deformation, which will cause target anchor points collection incompletely. So we introduce Algorithm~\ref{alg: TPS} to cover this situation.

\begin{algorithm}
\caption{\textcolor{black}{Constructing TPS transformation}}
\begin{algorithmic}[1]
\State \textbf{Input:} Given original image $\mathbf x_1$ (frame 1) with $r_1 \times c_1$  anchor points, each of which has coordinate   $\mathbf p^{(1)} [i,j]$, where $i  \in [ r_1 ]$,  $j   \in [ c_1 ]$ and $[n]$ denotes the integer set $\{ 1,2,\ldots, n\}$, target image  $\mathbf x_2$ (frame 2) with $r_2 \times c_2$  anchor points, each of which has coordinate   $\mathbf p^{(2)} [i,j]$, where $i \in  [r_2]$ and  $j  \in [ c_2]$,   distance tolerance   $\epsilon > 0$, and empty vectors $\tilde{\mathbf p}^{(1)} $ and $\tilde{\mathbf p}^{(2)} $.
% Anchor points' coordinates $\phi ^{ c_A \times r_A}$, target anchor points' coordinates $\psi^{ c_T \times r_T}$. 
%($c$ and $r$ means the column number and row number respectively.)
\State\textbf{Output:} Matched $r \times c$ anchor points $\tilde{\mathbf p}^{(1)}[i,j]$ versus  $\tilde{\mathbf p}^{(2)}[i,j]$ for $i \in [r]$ and $j \in [c]$, and TPS transformation $t_{\mathrm{TPS}}$ from $\mathbf x_1$ to $\mathbf x_2$.
%Transformation $t_{\mathrm{TPS}}$ that all anchor points in $\psi$ are transferred from $\phi$ in correct mapping.
%\State $\phi_0 \leftarrow \emptyset$
% \If {$c_1 = c_1 $ and $r_2 = r_2$}
%   %  \State $\phi_0  \leftarrow \phi$
%   % \State mapping $\phi$ and $\psi$ sequentially.
%   \State 
% \Else
 \For{$(i,j) \in [r_1] \times [c_1] $}
 \State given $\mathbf p^{(1)} [i,j]$ in $\mathbf x_1$, find the candidate of matching point $\mathbf p^{(2)} [i^\prime,j^\prime]$ by   nearest neighbor in $\mathbf x_2$,
 %$(i^\prime, j^\prime) \in  [r_2] \times [c_2]$
 \If{$\| \mathbf p^{(1)} [i,j] -  \mathbf p^{(2)} [i^\prime,j^\prime] \|_2 \leq \epsilon$} 
 \State matching $\mathbf p^{(1)} [i,j]$ with $\mathbf p^{(2)} [i^\prime,j^\prime]$, and adding them into  $\tilde{\mathbf p}^{(1)} $ and $\tilde{\mathbf p}^{(2)} $ respectively,
 \EndIf
 \EndFor
%     \LineComment $\psi$ is incomplete
%     \State $c, r$ $\leftarrow$ mapping $\phi$ and $\psi[0,0]$ by nearest neighbour.
%     \State $\phi_0 \leftarrow \phi[c:c+c_T,r:r+r_T]$
% % \EndIf
\State build TPS transformation $t_{\mathrm{TPS}}$ by solving Eq.~\eqref{eq: TPS_form} given $\tilde{\mathbf p}^{(1)} $ and $\tilde{\mathbf p}^{(2)} $.
%\State \textbf{return:} $T_{tps}$
\end{algorithmic}\label{alg: TPS}
\end{algorithm}

\section{Color transformation}\label{app: color}

\begin{figure}[h!]
     \begin{center}
     \begin{tabular}{ c c c}
    \includegraphics[width = 0.1\textwidth]{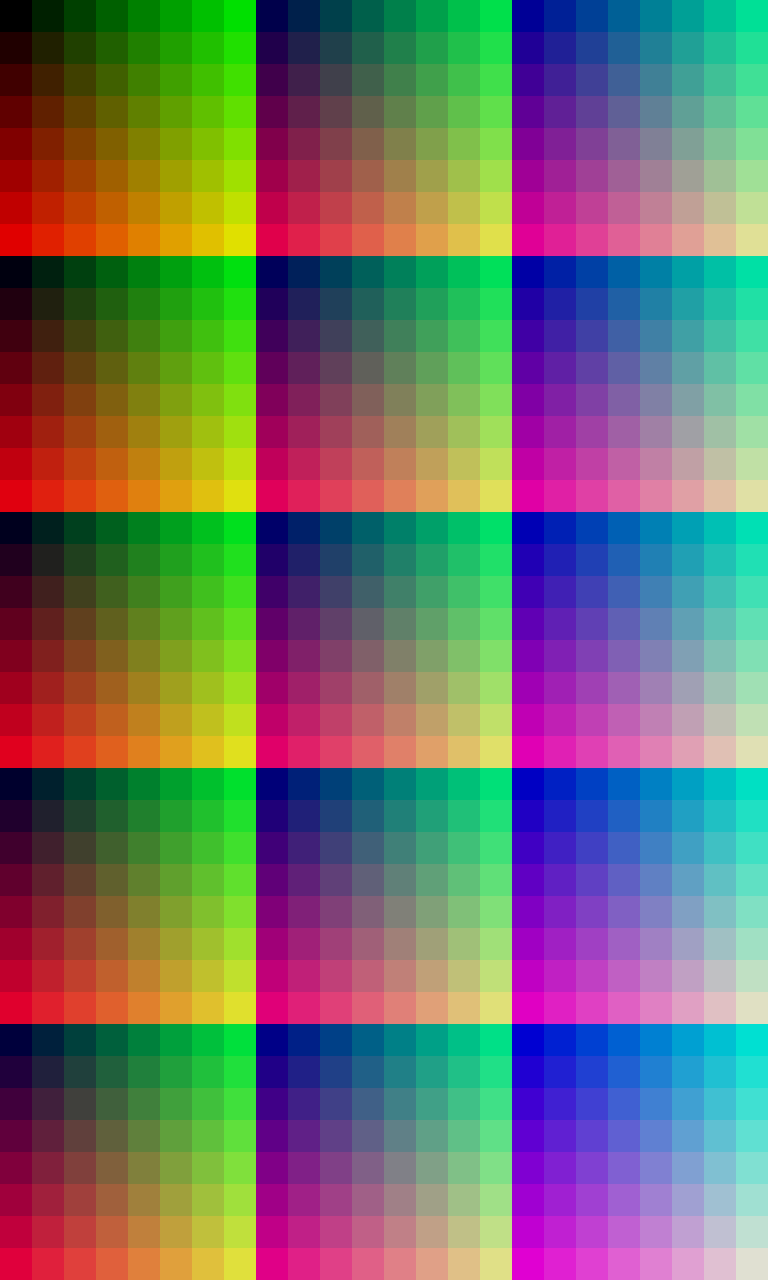}&
   \includegraphics[width = 0.1\textwidth]{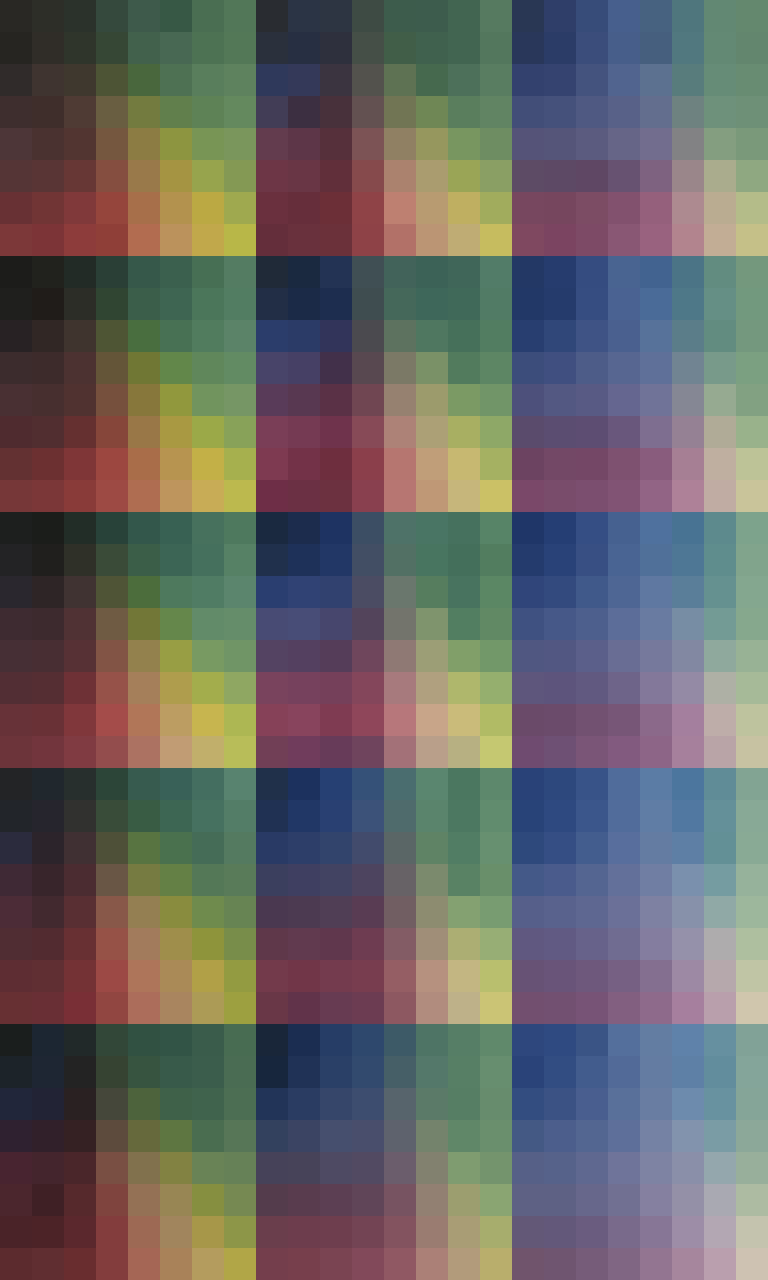}&
    \includegraphics[width = 0.1\textwidth]{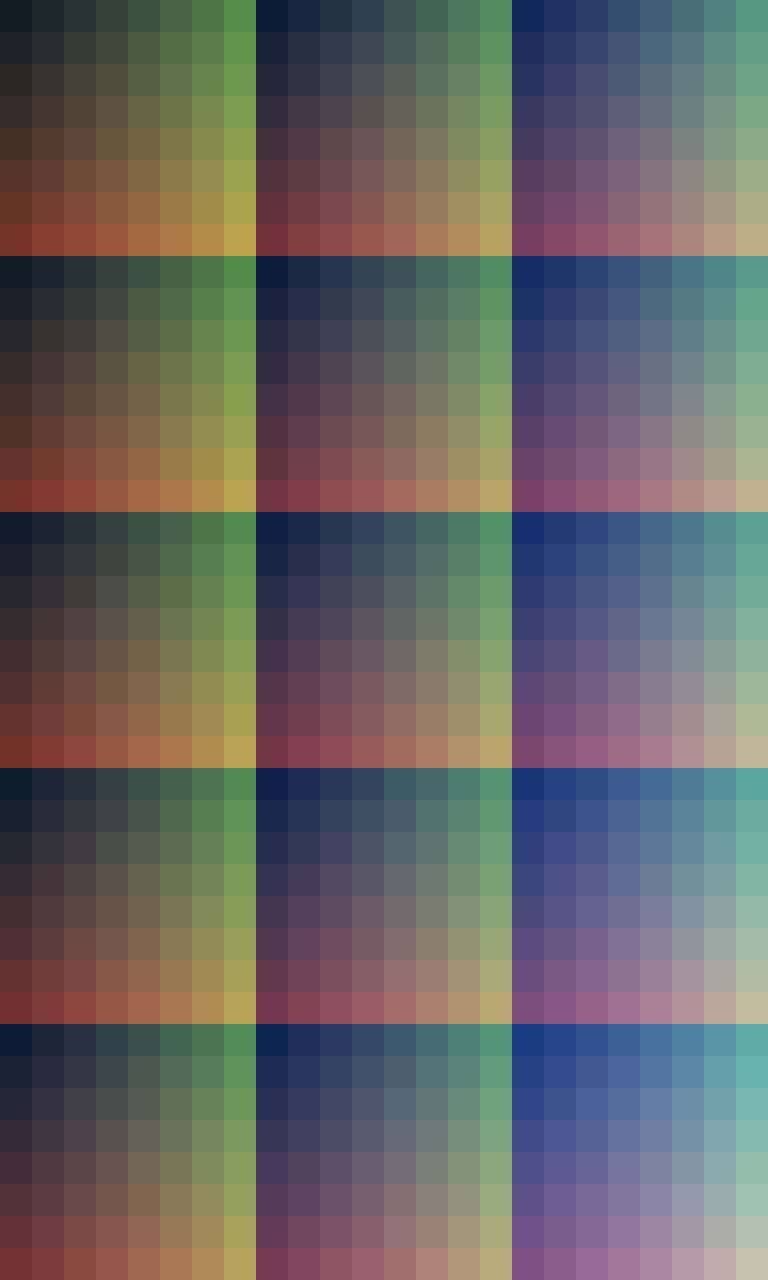}\\
    (a) &(b)&(c)
      \end{tabular}
      \vspace*{-2mm}
      \caption{Physical color transformation. (a): The digital color map (b): The  printed color map   on a T-shirt (captured by {the camera of iPhone X}).  (c): The predicted transformation  from (a) via the learnt polynomial regression.
      } 
      \label{tab: colormap}
      \end{center}
      \end{figure}

As shown in Figure\,\ref{tab: colormap}, we generate  the training dataset to map a digital color palette to the same one printed on a T-shirt. With the aid of $960$ color cell pairs. We learn the weights of the  quadratic polynomial regression by  minimizing the mean squared error of the predicted physical color (with the digital color in Figure\,\ref{tab: colormap}(a) as input) and the ground-truth physical color provided in Figure\,\ref{tab: colormap}(b). Once the color transformer $t_{\mathrm{color}}$ is learnt, we then incorporate it into \eqref{eq: perturbation_robust}.
% a multilayer perception (MLP) of $2$ hidden layers, each of which contains $256$ and $512$ neurons. Both the input and the output layers have dimension of $3$. As shown in Figure\,\ref{tab: colormap}, we generate  the training dataset to map a digital color palette to the same one printed on a T-shirt. With the aid of $960$ color cell pairs, we learn the weights of MLP by minimizing the mean squared error of the predicted physical color (with the digital color in Figure\,\ref{tab: colormap}(a) as input) and the ground-truth physical color provided in Figure\,\ref{tab: colormap}(b). 
% Once the MLP-based color transformer $t_{\mathrm{color}}$ is learnt, we then incorporate it into \eqref{eq: perturbation_robust}.
%Define $t_{\mathrm{color}}$.

% \end{myremark}

      \vspace*{-2mm}

\section{Formulation of attack loss}\label{app: loss}
\textcolor{black}{There are two possible options to formulate the attack loss $f$ to fool person detectors. First, $f$ is specified as the
misclassification loss, 
 commonly-used  in most of previous works. The goal is to misclassify the class `person' to any other incorrect class. 
 However, our work consider a more advanced disappearance attack, which enforces  the detector even not to draw the bounding box of the object `person'.
% In order to build the attack loss $f$, we  have two strategies: 1) misclassification attack (untarget attack, like attacking the  class `person' to any other classes such as `bottle' or `bicycle'). 2) disappearing attack: not only enforce the detector does not draw bounding box of the person but also does not recognize the person as any other classes.
For YOLOv2, we minimize the confidence score of all bounding boxes corresponding to the   class `person'. 
For Faster R-CNN, we attack all bounding boxes towards the class `background'. 
%this is a special target attack and harder than conventional untarget attack. 
% So we minimize this confidence score which greater than a threshold. I n conclusion, we define our attack loss as:
Let $\mathbf x_i^{\prime}$ be a perturbed video frame, the attack loss in (6) is then given by
\begin{align}\label{eq: attack_loss}
    \begin{array}{cl}
\displaystyle       f( \mathbf x_i^{\prime} ) = 
  \max_j \{  \max \{  p_j(\mathbf x_i^\prime) , \nu \} \cdot \mathbbm{1}_{ | B_j \cap M_{p,i} | > \eta } \},
%  max_\mathbf{b}\{ max \{\mathbf{P}, v\} \circ \mathbbm{1}_{\mathbf{b}, M_{p,i}}\}
%\\
    %\st      &  \boldsymbol{\delta} \in \mathcal C,
    \end{array}
\end{align}
where 
$p_j(\mathbf x_i^\prime) $ denotes the confidence score of the $j$th bounding box  for YOLOv2 or  the probability of the `person' class at the $j$th bounding box  for Faster R-CNN, $\nu$ is a confidence threshold, the use of $\max \{  p_j(\mathbf x_i^\prime) , \nu \}$  enforces the optimizer to minimize the bounding boxes of high probability (greater than $\nu$), $B_j$ is the $j$th bounding box, $M_{p,i}$ is the known bounding box encoding the person's region, the quantity $| B_j \cap M_{p,i} |  $ represents the intersection between $B_j$ and $M_{p,i}$, $|\cdot |$ is the cardinality function,  and
$\mathbbm{1}_{ | B_j \cap M_{p,i} | > \eta }$ is the indicator function, which returns $1$ if $B_j$ has at least $\eta$-overlapping with $M_{c,i}$, and $0$ otherwise. 
In  Eq.\eqref{eq: attack_loss}, the quantity $\max \{  p_j(\mathbf x_i^\prime) , \nu \} \cdot \mathbbm{1}_{ | B_j \cap M_{p,i} | > \eta }$ characterizes the bounding box of our interest with both high probability and  large overlapping with $M_{p,i}$.
And the eventual loss in  Eq.\eqref{eq: attack_loss} gives the largest probability for detecting a     bounding box of the object `person'. 
}
% $\theta$ is the parameters of given object detector, $\mathbf{P}$ is the probabilities of bounding box existence for YOLOv2 or the `person' class for faster R-CNN of the set of bounding boxes $\mathbf{b}$. The attack loss is beneficial to $v$ as a threshold to concentrate on the worth bounding boxes which has higher $\mathbf{P}$. The indicator function $\mathbbm{1}$ help us filter out the bounding boxes which has very low overlap with our attack object $M_{p,i}$ by calculate their IOU scores.

%  \textcolor{blue}{PY: it's still not clear what transformations we actually use in the attack formulation. Please specify the transformations in our setting.}

% \textcolor{Quanfu}{
% \begin{align}\label{eq: attack_loss}
%     \begin{array}{cl}
% \displaystyle       f( \mathbf x_i^{\prime} ) = 
%   \max_{B_j \in B} \{  p(B_j) \},
% %  max_\mathbf{b}\{ max \{\mathbf{P}, v\} \circ \mathbbm{1}_{\mathbf{b}, M_{p,i}}\}
% %\\
%     %\st      &  \boldsymbol{\delta} \in \mathcal C,
%     \end{array}
% \end{align}
% where B is a set of proposals that significantly overlap with the ground-truth person wearing adv-T and have a confidence score greater than $v$ (see Fig 3 in the main paper).
% } 

\section{Hyperparameter setting}\label{app: hyper}
When solving Eq.~\eqref{eq: attack_single}, we use Adam optimizer~\cite{kingma2014adam} to train 5,000 epochs with the initial learning rate, $1 \times 10^{-2}$. The rate is decayed when the loss ceases to decrease. The regularization parameter $\lambda$ for total-variation norm is set as  $3$. In Eq.~\eqref{eq: attack_multiple},  we set $\gamma$ as 1, and solve the min-max problem by 6000 epochs with the initial learning rate $1 \times 10^{-2}$. In  Eq.~\eqref{eq: perturbation_robust}, the details of  transformations $t$ are shown in Table~\ref{table:trans_details}.

\begin{table}[htb]
% \vspace*{-5mm}
%\begin{table}[htb]
 \centering
\begin{tabular}{c|cc}
\toprule[1pt]
Transformation & Minimum & Maximum \\
\midrule
Scale & 0.5 & 2 \\
Brightness & -0.1 & 0.1 \\
Contrast & 0.8 & 1.2 \\
Random uniform noise & -0.1 & 0.1 \\
Blurring & \multicolumn{2}{c}{average pooling/filter size = 5} \\

\bottomrule[1pt]
\end{tabular}
 \caption{The conventional transformations $t$ in Eq.~\eqref{eq: perturbation_robust}. 
 %we used during training with their parameters setting. 
 }
  \label{table:trans_details}
%   \vspace*{-2mm}
\end{table}

% To generate our perturbation we use scaling, translation, brightness, blurring and random noise in addition to TPS transformation.

\begin{figure}[h!]
     \begin{center}
     \begin{tabular}{ c|ccc}
     \toprule
     $\lambda$ & 1 &3 &5\\ \hline
    &\includegraphics[width = 0.1\textwidth]{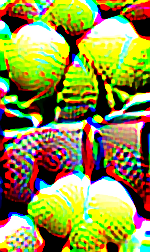}&
    \includegraphics[width = 0.1\textwidth]{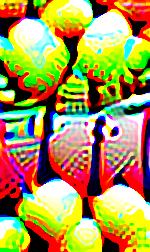}&
    \includegraphics[width = 0.1\textwidth]{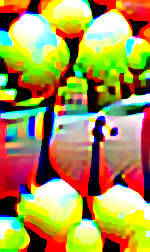}\\
    \hline
    digital & 82\% & 74\% & 69\% \\ \hline
   physical & 51\% &57\% &55\%\\
   \bottomrule
      \end{tabular}
      \vspace{2mm}
      \caption{{ASR   v.s.   $\lambda$ against YOLOv2.
      }}
      \label{fig: lambda}
      \end{center}
      \end{figure}

\textcolor{black}{In experiments, we find that the hyperparameter $\lambda$ strikes a balance between the fine-gained perturbation pattern and its smoothness. 
%trade-off between the digital world and physical world performance.
As we can see in Figure~\ref{fig: lambda}, when $\lambda$ is smallest (namely, $\lambda = 1$), the perturbation can achieve the best ASR (82\% ) against YOLOv2 in the digital space, however when we test the digital pattern in the physical world, the attacking performance  drops to 51\% (worse than the case of $\lambda = 3$) as the non-smooth (sharp) perturbation pattern might not be well captured by a real-world camera. In our experiments, we choose $\lambda = 3$ for the best tradeoff between digital and physical results.}
% so that the physical performance even drop to 51\%, which worse than 57\% when $\lambda=3$. So we set $\lambda$ as $3$ empirically.

\begin{figure}[h!]
     \begin{center}
     \begin{tabular}{c}
    \includegraphics[width = 0.4\textwidth]{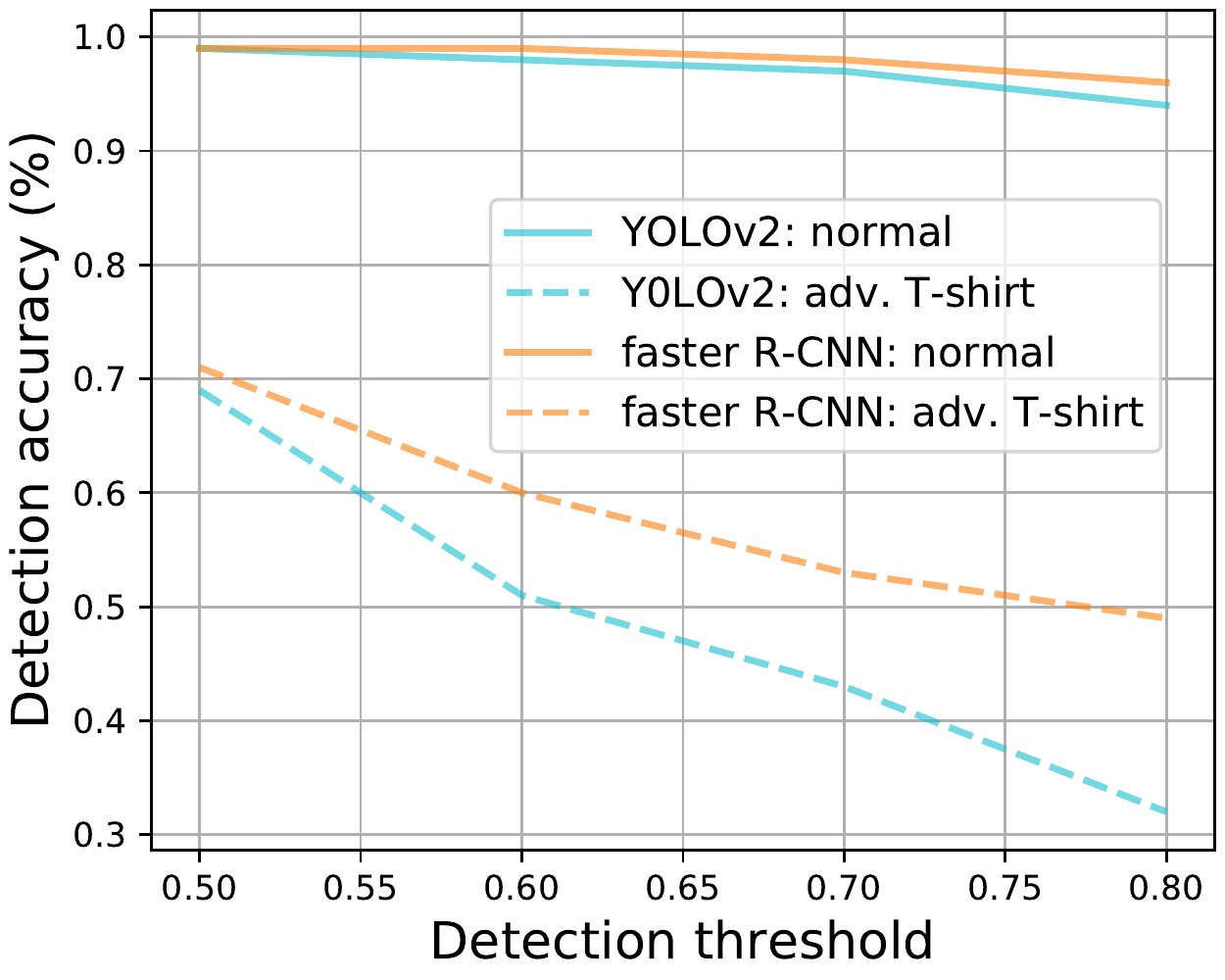}
      \end{tabular}
    %   \vspace{2mm}
      \caption{{The detection accuracy of YOLOV2 and Faster R-CNN under different detection thresholds . `Normal' means the case of   persons wearing  normal clothing, and `adv. T-shirt' means the case of   persons wearing the adversarial T-shirt.
      }}
      \label{fig: threshold}
      \end{center}
      \end{figure}

\textcolor{black}{For a real-world deployment of a person detector, the minimum detection threshold needs to be empirically determined to obtain a good tradeoff between detection accuracy and false alarm rates. In our physical-word testing, we set the threshold  to 0.7 for Faster R-CNN and YOLOv2, at which both of them achieve detection accuracy over 97\% on person wearing normal clothing. The sensitivity analysis of this threshold is provided in Figure~\ref{fig: threshold}.}      
%\textacolor{black}{In the physical-world testing, we set the minimum detection threshold  as 0.7 for both Faster R-CNN and YOLOv2. We also evaluate the sensitivity of this threshold  in Figure~\ref{fig: threshold}.} 
% As we can see, the detection recall   drops as the detection threshold growing. So we select a trade-off point as $0.7$ that the recall of person worn normal clothes hold high performance.

\section{Dataset details}\label{app:exp_setup}

In Table\,\ref{table:dataset}, we summarize dataset we used in Section \ref{exp: digital} and \ref{exp:physical}.

\begin{table}
    \caption{Summary of our collected dataset in each scenes. The values in the table are presented by number of videos (total number of frames) in each scene, ie, 4 (177) means 4 videos and 177 frames in total.
    %Note that in test(physical), all dataset are recollected at seen or unseen scenes. 
    }
    \label{table:dataset}
    \centering
    \resizebox{0.9\linewidth}{!}{
    \begin{tabular}{c|ccc|cc|c}
        \toprule
        \multirow{2}{*}{videos (frames)} &\multicolumn{3}{c|}{indoor} & \multicolumn{2}{c|}{outdoor} & overall\\
        \cmidrule{2-6}
                 &    office  & elevator  &  hallway  & street1 & street2 & \\
        \midrule[1pt]
     single-person   & 4 (177) & 4 (135) & 4 (230) & 4 (225) & 4 (240) & 20 (1007) \\
     multi-persons  & 4 (162) & 4 (132) & 4 (245) & 4 (230) & 4 (227)  & 20 (996)\\
        \midrule
     train           & 6 (245) & 6 (180) & 6 (335) & 6 (344) & 6 (365) & 30 (1469) \\
     test (digital)   & 2 (94) & 2 (87) & 2 (140) & 2 (111) & 2 (102) & 10 (534) \\
     \midrule[1pt]
                 &    unseen  & elevator  &  hallway   &\multicolumn{2}{c|}{street3}& \\
     \midrule
     test (physical)   & 6 (236) & 6 (184) & 6 (220)  &\multicolumn{2}{c|}{6 (288)} & 24 (928) \\
        \bottomrule
    \end{tabular}
    }
\end{table}

In Section \ref{sec:abalation} for ablation study on parameter sensitivity and generalization to more complex testing scenarios, 
%, in order to analyse the sensitivity of each controllable variable, 
we further collected some new  test data. Specifically, we considered the scenario of five people (two females and three males) for ablation  study and none of them appeared in the original training and testing datasets. We recorded   multiple videos by using two cameras (one iPhone X and one iPhone XI) and reported the resulting  ASR in average. 
%In contrast with the previous physical-world evaluation, these experiments (except the distance one) have all the actors stay at a fixed location, usually 1-3m away from the camera. 

\section{More experimental results}\label{app:exp}

\begin{figure*}[htb]
   \centering
   \adjustbox{max width=1\textwidth}{
\hspace*{-0.1in}\begin{tabular}{p{0.22in}p{0.65in}p{0.65in}p{0.65in}p{0.65in}p{0.65in}p{0.65in}p{0.65in}}

%  \begin{tabular}[l]{@{}l@{}}
 %\rotatebox{90}{ \footnotesize{Africa elephant}}  
 %\rotatebox{90}{\parbox{0.9in}{\centering Africa elephant}}
 \hspace*{0.05in} \rotatebox{90}{\parbox{1.2in}{\centering \footnotesize \TPS}}
 &   \hspace*{-0.1in} 
\includegraphics[width=0.7in]{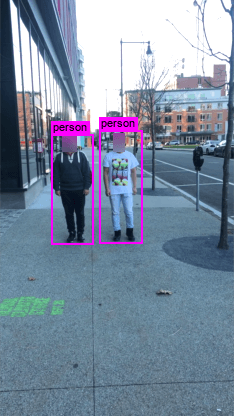}& \hspace*{-0.1in}  
\includegraphics[width=0.7in]{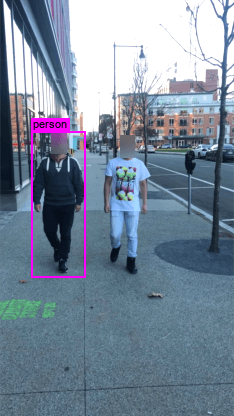}& \hspace*{-0.1in}  \includegraphics[width=0.7in]{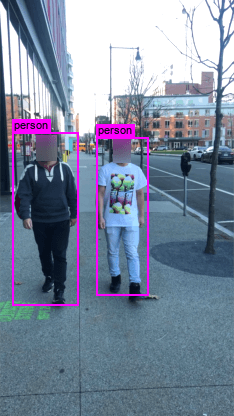}& \hspace*{-0.1in}  \includegraphics[width=0.7in]{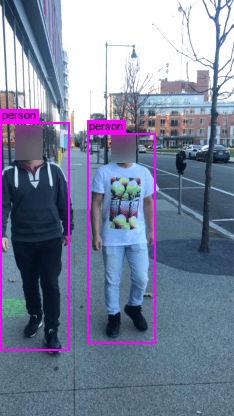}& \hspace*{-0.1in}  \includegraphics[width=0.7in]{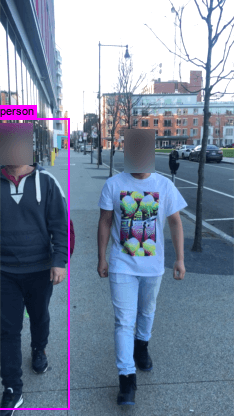}& \hspace*{-0.1in}  \includegraphics[width=0.7in]{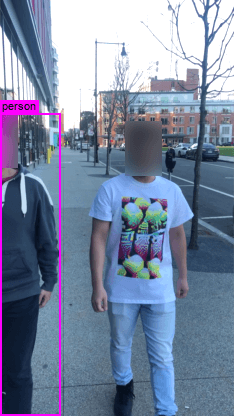}& \hspace*{-0.1in}   \includegraphics[width=0.7in]{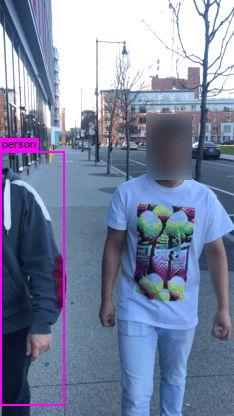}
\\

 \hspace*{0.05in} \rotatebox{90}{\parbox{1.2in}{\centering \footnotesize \affine}}
 &   \hspace*{-0.1in} 
\includegraphics[width=0.7in]{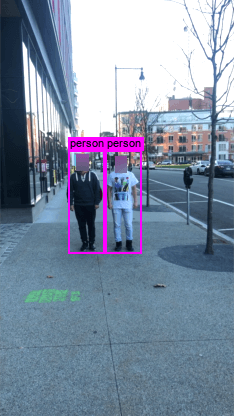}& \hspace*{-0.1in}  
\includegraphics[width=0.7in]{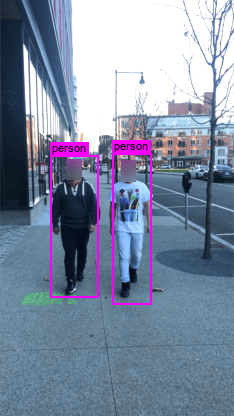}& \hspace*{-0.1in}  \includegraphics[width=0.7in]{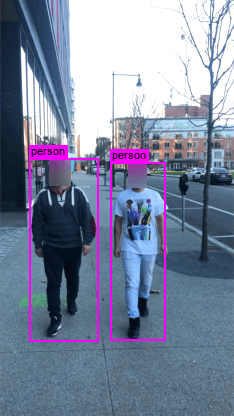}& \hspace*{-0.1in}  \includegraphics[width=0.7in]{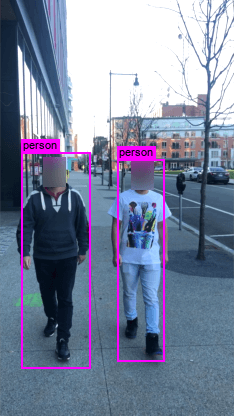}& \hspace*{-0.1in}  \includegraphics[width=0.7in]{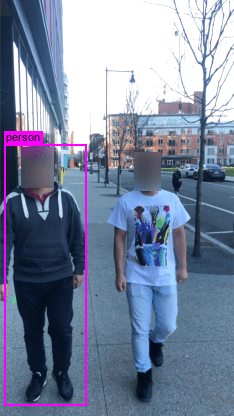}& \hspace*{-0.1in}  \includegraphics[width=0.7in]{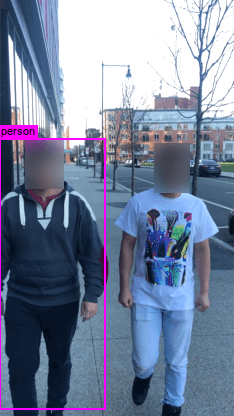}& \hspace*{-0.1in}   \includegraphics[width=0.7in]{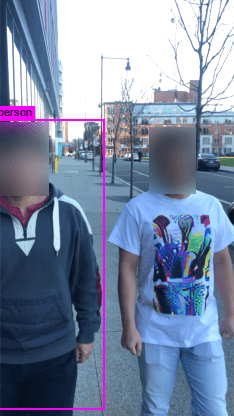}
\\

 \hspace*{0.05in} \rotatebox{90}{\parbox{1.2in}{\centering \footnotesize \baseline}}
 &   \hspace*{-0.1in} 
\includegraphics[width=0.7in]{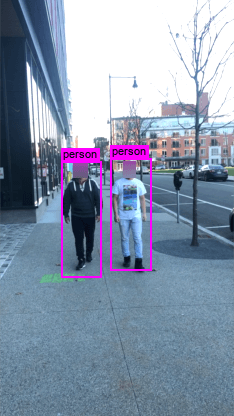}& \hspace*{-0.1in}  
\includegraphics[width=0.7in]{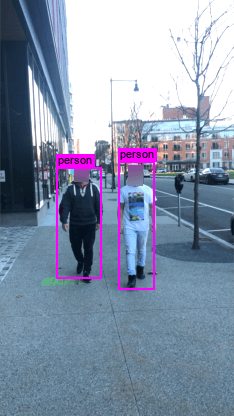}& \hspace*{-0.1in} \includegraphics[width=0.7in]{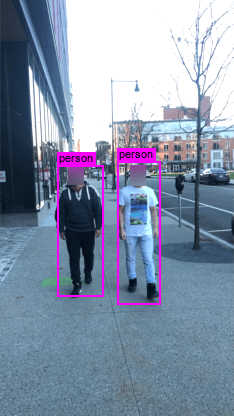}& \hspace*{-0.1in} \includegraphics[width=0.7in]{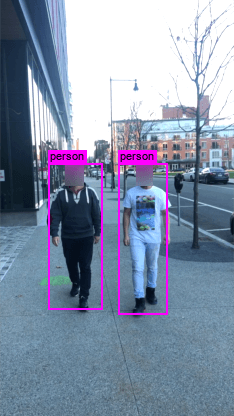}& \hspace*{-0.1in} \includegraphics[width=0.7in]{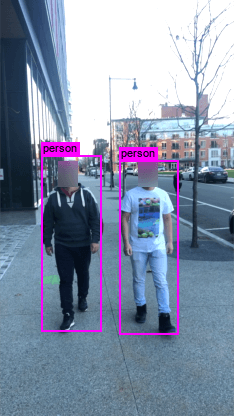}& \hspace*{-0.1in} \includegraphics[width=0.7in]{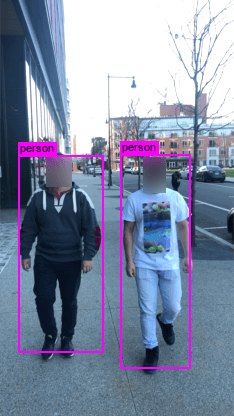}& \hspace*{-0.1in}  \includegraphics[width=0.7in]{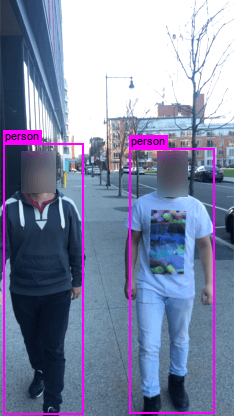}
\\

 \hspace*{0.05in} \rotatebox{90}{\parbox{1.2in}{\centering \footnotesize \TPS}}
 &   \hspace*{-0.1in} 
\includegraphics[width=0.7in]{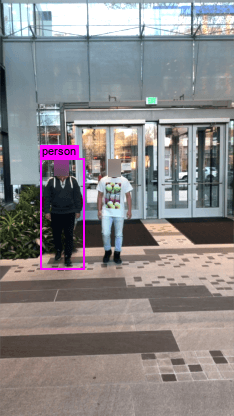}& \hspace*{-0.1in}  
\includegraphics[width=0.7in]{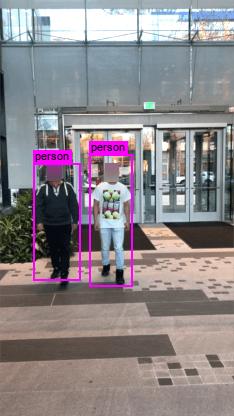}& \hspace*{-0.1in}  
\includegraphics[width=0.7in]{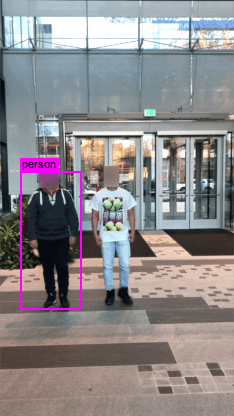}& \hspace*{-0.1in}  
\includegraphics[width=0.7in]{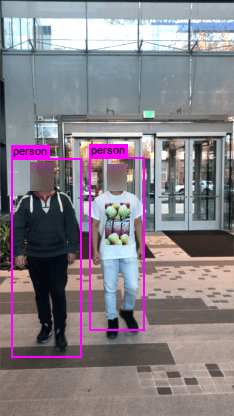}& \hspace*{-0.1in}  
\includegraphics[width=0.7in]{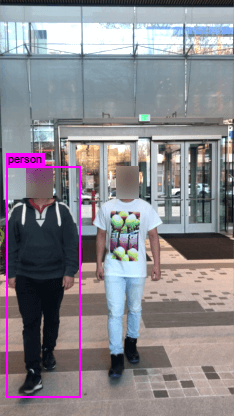}& \hspace*{-0.1in}  
\includegraphics[width=0.7in]{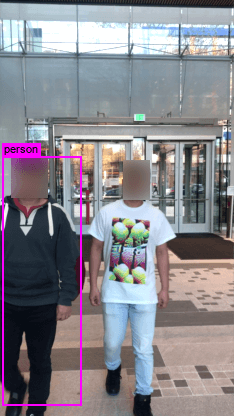}& \hspace*{-0.1in}  
\includegraphics[width=0.7in]{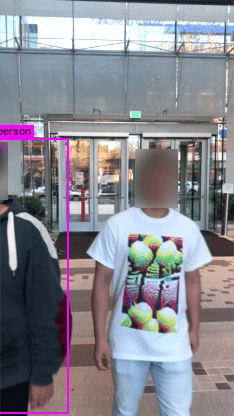}
\\

 \hspace*{0.05in} \rotatebox{90}{\parbox{1.2in}{\centering \footnotesize \affine}}
 &   \hspace*{-0.1in} 
\includegraphics[width=0.7in]{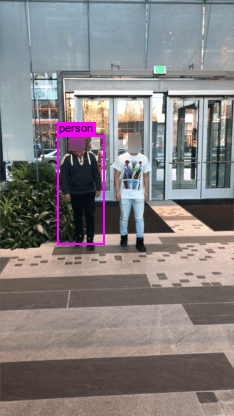}& \hspace*{-0.1in}  
\includegraphics[width=0.7in]{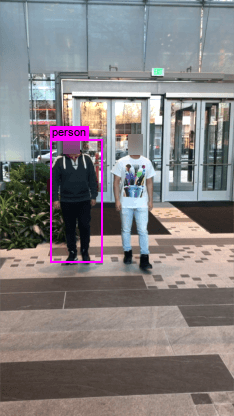}& \hspace*{-0.1in}  
\includegraphics[width=0.7in]{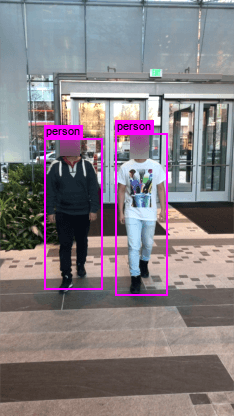}& \hspace*{-0.1in}  
\includegraphics[width=0.7in]{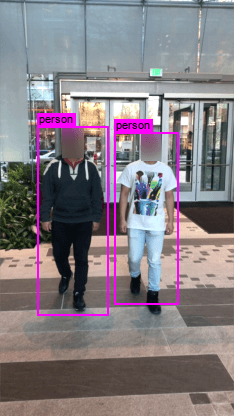}& \hspace*{-0.1in}  
\includegraphics[width=0.7in]{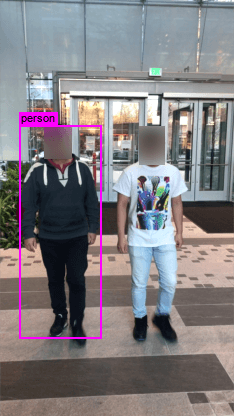}& \hspace*{-0.1in}  
\includegraphics[width=0.7in]{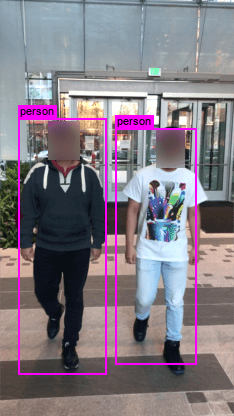}& \hspace*{-0.1in}  
\includegraphics[width=0.7in]{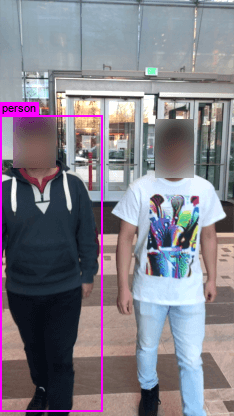}
\\

% & 
%  \hspace*{-0.1in} \parbox{0.65in}{\centering \footnotesize (a)} & \parbox{0.65in}{\centering \footnotesize (b)}
% & \parbox{0.65in}{\centering \footnotesize (c)} & \parbox{0.65in}{\centering \footnotesize (d)}
% & \parbox{0.65in}{\centering \footnotesize (e)} & \parbox{0.65in}{\centering \footnotesize (f)}
% \vspace{-6mm}
\end{tabular}
}
% \vspace{-2mm}
\caption{\footnotesize{
Some testing frames in the physical world using adversarial T-shirt against YOLOv2. All frames are performed by two persons with one wearing the proposed adversarial T-shirt, generated by our method (\TPS), \affine and \baseline. The first three rows: an unseen outdoor scenes. The last two rows: an unseen indoor scenes.
}}
    \label{fig: physical_examples}
    %\vspace*{-4mm}
\end{figure*}

In Figure \ref{fig: physical_examples}, we demonstrate  our  physical-world  attack results in two scenarios: a)     adversarial T-shirts generated by \TPS, \affine and \baseline  in an outdoor scenario (the first three rows), b)  adversarial T-shirts generated by \TPS and  \affine in an unseen scenario (at a location never seen in the training dataset).
As we can see,  our method outperforms affine and baseline. In the absence of TPS, 
adversarial T-shirts generated by affine and baseline   fail in most of cases, implying  the importance of  TPS to model the T-shirt deformation.
When a person whom wears the adversarial T-shirt walks towards   the camera, as expected, the detector also becomes easier to be attacked.

 \begin{figure*}[htb]
  \centering
\begin{tabular}{p{0.60in}p{0.60in}p{0.60in}|p{0.60in}p{0.60in}p{0.60in}}
 \multicolumn{3}{c|}{ \footnotesize \affine} & \multicolumn{3}{c}{ \footnotesize \TPS} 
\\
\includegraphics[width=0.60in]{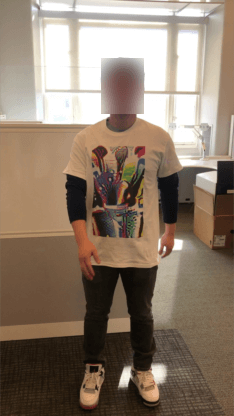}& 
\includegraphics[width=0.60in]{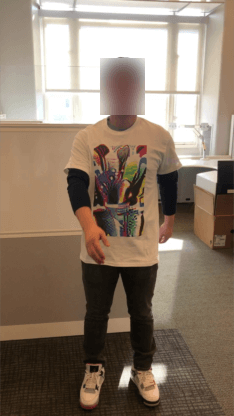}& 
\includegraphics[width=0.60in]{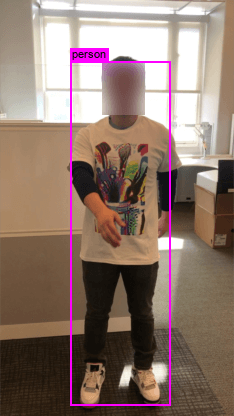}&
\includegraphics[width=0.60in]{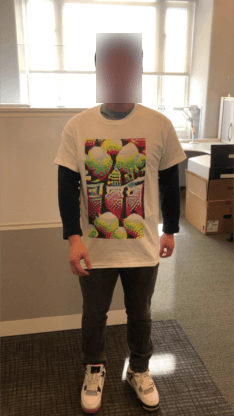}& 
\includegraphics[width=0.60in]{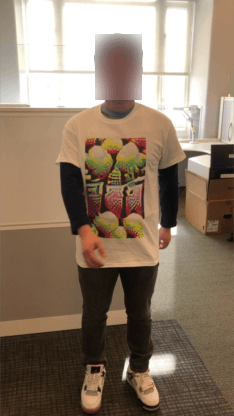}& 
\includegraphics[width=0.60in]{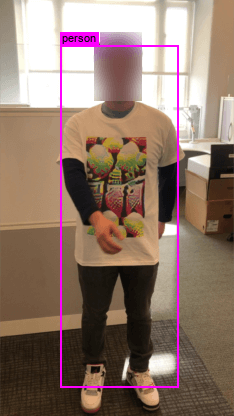}
\end{tabular}
% \vspace{-2mm}
\caption{\footnotesize{
When \affine and \TPS happen occlusion by hand.
}}
    \label{fig: occlusion}
\end{figure*}

% \end{document}